\documentclass[11pt]{article}

\usepackage[T1]{fontenc}
\usepackage[margin=1in]{geometry}
\usepackage{amsmath,amssymb}
\usepackage{graphicx}
\usepackage{tabularx}
\usepackage{caption}
\usepackage{float}
\usepackage{booktabs}
\usepackage{longtable}
\usepackage{array}
\usepackage{placeins}
\usepackage{microtype}
\usepackage{xcolor}
\usepackage{tikz}
\usetikzlibrary{arrows.meta,calc}
\usepackage[numbers,sort&compress]{natbib}
\usepackage[hidelinks]{hyperref}
\hypersetup{pdftitle={Causal understanding of agentic capabilities via randomized interventions}}
\graphicspath{{figures/}}


\newcommand{\E}{\mathbb{E}}
\newcommand{\ind}{\mathbf{1}}
\newcommand{\argmin}{\operatorname*{arg\,min}}
\newcommand{\feat}[1]{\texttt{\detokenize{#1}}}
\definecolor{tsk}{HTML}{5E8DB8}
\definecolor{sta}{HTML}{D9B65B}
\definecolor{trt}{HTML}{B8472A}
\definecolor{out}{HTML}{9C7F66}
\definecolor{ink}{HTML}{3B4652}
\definecolor{mute}{HTML}{9AA3AD}

\usepackage{color-edits}
\addauthor{vs}{red}

\title{Knowing When to Intervene: Causal Evaluation of AI Agents}

\author{
   Ayush Sawarni\footnote{Corresponding author}, Wenqian Xing, Tijana Zrnic, Ramesh Johari, and Vasilis Syrgkanis\\[0.5em]
  Stanford University\\[0.2em]
  \texttt{\{ayushsaw,wxing,tijana.zrnic,rjohari,vsyrgk\}@stanford.edu}
}
\hypersetup{pdfauthor={Ayush Sawarni, Wenqian Xing, Tijana Zrnic, Ramesh Johari, Vasilis Syrgkanis}}
\date{}

\begin{document}
\maketitle
% \ayush{
% Other candidate titles:
% \begin{itemize}
%     \item When Should I Verify What My Agent Did? Causal Evaluation of Agentic AI via Randomized Interventions
%     \item Causal Understanding of Agentic Capabilities via Randomized Interventions
%     \item Knowing When to Intervene: Causal Evaluation of AI Agents
%     \item Knowing When to Intervene: Causal Evaluation of AI Agents via Sequential Randomization
% \end{itemize}
% }

\begin{abstract}
Current evaluations of agentic AI systems primarily rely on benchmarks. However, the static nature of benchmarks makes them uninformative about the value of auditing interventions at deployment time.
% These benchmarks consist of many long-horizon tasks, along with the environments, tools, and context available to agents, as well as criteria for evaluating their current trajectories during task execution.
% While a valuable source of data for characterizing agent behavior, benchmarks have significant limitations in guiding the use of agentic AI in real-world settings.
% Current evaluations of agentic AI typically rely on benchmarks that includes long-horizon tasks, interactive environments, available tools, and criteria for assessing agent trajectories. While these benchmarks provide valuable insights into agent behavior, they offer limited guidance on how interventions during execution might affect downstream behavior and final task outcomes.
% In real world use of agentic AI, humans are often determining within a trajectory whether or not to intervene, e.g., by verifying AI agent output, or by steering the agent to a different behavior.  Whether or not to apply such an intervention is a fundamentally causal inferential question: the user is ascertaining whether the causal effect of such an intervention is beneficial on the final outcome.  This causal effect is particularly challenging because it is a dynamic target: the impact of an intervention depends on the current state of context and environment, and the future behavior of the agentic policy.  Benchmarks as currently constituted are incapable of providing any insight into these causal effects.
For reliable deployment, we often need to determine {\em within an agent's trajectory} whether or not to intervene, e.g., by verifying the agent's current output or steering it toward different behavior, using either human review or more costly models or time-consuming agentic supervision.
The value of such interventions is fundamentally a {\em causal inference} question, asking whether the {\em causal effect} of such an intervention is beneficial for the final outcome.
Estimating this causal effect is particularly challenging due to the dynamic nature of the problem: the impact of an intervention depends on the current context and environment, as well as the agent's future behavior.
% Benchmarks, as currently designed, are incapable of providing insight into these causal effects.
We develop the first methodologically grounded framework for evaluating the causal effects of randomized interventions within agent trajectories.
By leveraging a relatively small dataset of agent trajectories collected under randomized interventions such as verification or screening, we can estimate the dynamic causal effect of a rich class of (contextual) intervention policies. The structural parameters that we estimate can also be used to construct safety leaderboards that capture the risk of various agentic actions, which can be used by practitioners for audit prioritization.
% Remarkably, despite the simplicity of the randomization scheme, we show that these data can be used to credibly estimate the causal effects of a rich class of intervention policies, including those that depend on the agent's observable history.
Applying our methods to $\tau$-bench Retail, ToolSandbox, and ImpossibleBench \citep{yao2025tau,barres2025tau2,lu2025toolsandbox,zhong2025impossiblebench}, we find that intervention effects vary substantially across agent actions and states.
More broadly, our work suggests a rich research agenda at the intersection of causal inference and agentic evaluation, which we believe is essential for the reliable and safe deployment of AI agents in the real world.\footnote{We used ChatGPT, Codex, Claude, and Claude Code to assist with writing, implementing experiments, and analyzing results. The authors take responsibility for the manuscript and results.}

\end{abstract}

\section{Introduction}
\label{sec:introduction}

% Agentic AI systems are designed to complete complex, long-horizon tasks through sequences of actions and tool calls in interactive environments. Evaluating these systems typically involves benchmarks that specify tasks, environments, available tools, and criteria for assessing agent performance. For example, in $\tau^2$-bench, a customer-support agent interacts with simulated customers to collect account information, update order statuses through external tools, and respond to customer requests \citep{barres2025tau2,lu2025toolsandbox}.
% % In such settings, task outcomes depend on the entire state-action trajectory generated through interactions between the agent and its environment, rather than on a single model-generated response.
% In such settings, task outcomes depend on the entire state-action trajectory generated through interactions between the agent and the environment.

Agentic AI systems are designed to solve complex, long-horizon tasks through a sequence of decisions and tool calls in interactive environments.
Current evaluations of these systems typically rely on benchmarks, each of which fixes a set of tasks, an environment with its tools and instructions, and criteria for scoring the outcome \citep{jimenez2024swebench,zhou2024webarena,liu2024agentbench,yao2025tau,barres2025tau2}.
For example, in $\tau$-bench Retail \citep{yao2025tau}, a customer-support agent interacts with simulated customers, retrieves account information, and manages orders through external tools.
The task outcomes then depend on the entire trajectory of interactions between the agent and the environment.
While a valuable source of data for characterizing agent capabilities and safety, benchmarks have significant limitations in guiding how we intervene during real-world use.

% However, existing benchmarks primarily characterize agent behavior and performance under a given evaluation setting, rather than how interventions during execution might affect downstream outcomes \citep{jimenez2024swebench,zhou2024webarena,liu2024agentbench,yao2025tau,barres2025tau2,trivedi2024appworld,debenedetti2024agentdojo,zhong2025impossiblebench}.
% In real-world deployments, humans often face decisions about whether to intervene at intermediate points in an agent's trajectory, such as verifying a proposed tool call, providing corrective feedback, or screening potentially adversarial content.
% For example, would reviewing a proposed database write prevent a task failure, or would withholding an adversarial message prevent an agent from violating its instructions?
% These decisions raise fundamentally causal questions: whether an intervention improves the final outcome depends on the agent's current context and environment, as well as how the intervention affects its subsequent behavior.
% While existing benchmarks provide valuable information about agent capabilities and safety, they do not directly identify these dynamic causal effects.

In real-world deployments, humans need to determine {\em within an agent's trajectory} whether or not to intervene, e.g., by verifying a proposed tool call, steering the agent toward a different action, or screening potentially adversarial content. This review process can correspond to either human review in high-stakes domains, or, in medium-stakes domains, to more costly (and more powerful) model supervision, or supervision via a more time-consuming agentic harness.
% For instances,
% Would reviewing a proposed database write prevent a task failure?
% Would withholding an adversarial message prevent an agent from violating its instructions?
% Such decisions are fundamentally {\em causal inference} questions: the user must determine whether applying an intervention would improve the final outcome.
For instance, would reviewing a proposed database write prevent a task failure, or would withholding an adversarial message prevent an agent from violating its instructions?
Such decisions are fundamentally {\em causal inference} questions: the user must determine whether applying an intervention would improve the final outcome.

% These causal questions cannot be reliably answered by analyzing observed agent trajectories alone. While failed trajectories may reveal potential intervention points, an apparent mistake may be recoverable or unrelated to the eventual failure. Moreover, comparisons across observed trajectories can be confounded by the agent's evolving history. For example, agents receiving assistance after encountering errors may exhibit lower success rates simply because they face more difficult situations, rather than because the assistance is ineffective. 
% The dynamic environment also introduces another complexity: intervening at one decision can alter system states, which in turn affect the agent's future actions and intervention opportunities.
% Consequently, the causal effect of an intervention depends not only on where it is applied, but also on how the rest of the agent's trajectory unfolds.

Estimating this causal effect is particularly challenging because it is a {\em dynamic treatment effect} problem.
The impact of an intervention during task execution depends on the current context and environment, as well as the agent's future behavior under its policy.
In particular, an intervention can change subsequent states, actions, and intervention opportunities, so its effect also depends on how later interventions are assigned.
% Passively observed trajectories from benchmarks or trajectory-level randomized experiments (e.g., \cite{arena2026agentarena,camuffo2026beyond,lu2026agent}) do not fully resolve these questions.
Neither observed benchmark trajectories nor data from trajectory-level randomized experiments (see, e.g., \cite{arena2026agentarena,camuffo2026beyond,lu2026agent}) fully resolve these questions.
% An apparent mistake may be recoverable or unrelated to the eventual failure, and comparisons between assisted and unassisted trajectories can be confounded by the agent's evolving history.
% For example, agents receiving help after errors may have lower success rates because they face more difficult situations, rather than because assistance is ineffective.
While analyses of observed benchmark trajectories can reveal potential intervention points, they cannot by themselves identify the causal effects of intervening at those points.

% In principle, a fully controllable simulator could compare interventions by restoring an agent to a checkpoint and running different continuations.
% However, repeating these comparisons across many decisions can be expensive, and real-world interactions may not be replayable.
% An episode-level A/B test could instead compare complete intervention policies, such as always intervening versus never intervening, but each such comparison addresses only the selected policies.
% Even with $m$ fixed binary intervention opportunities, there are $2^m$ possible intervention schedules, before considering policies that adapt to the observed history.
% A more flexible design is needed to study when interventions help across the range of situations an agent encounters.

\begin{figure}[t]
\centering
\begin{tikzpicture}[
  font=\footnotesize, >={Stealth[length=1.8mm]},
  lab/.style={font=\footnotesize\bfseries, text=ink},
  card/.style={draw=ink!45, rounded corners=2pt, minimum width=25mm, minimum height=7mm,
               align=center, inner sep=2pt, fill=white, font=\scriptsize},
  prop/.style={card, fill=sta!16, draw=sta!80!black, font=\ttfamily\scriptsize},
  help/.style={card, minimum height=6mm, fill=trt!10, draw=trt!60},
  skip/.style={card, minimum height=6mm, fill=ink!5},
  coin/.style={circle, line width=.5pt, minimum size=5.2mm, inner sep=0pt, font=\scriptsize\bfseries},
  heads/.style={coin, fill=trt, text=white, draw=trt!70!black},
  tails/.style={coin, fill=white, text=ink, draw=ink!70},
  flow/.style={->, draw=ink!65, line width=.6pt},
  st/.style={circle, fill=sta, draw=sta!65!black, minimum size=8.6mm, inner sep=0pt, font=\small},
  tr/.style={circle, fill=trt, draw=trt!65!black, text=white, minimum size=8.6mm, inner sep=0pt, font=\small},
  ou/.style={circle, fill=out, draw=out!65!black, text=white, minimum size=9mm, inner sep=0pt, font=\small},
  tk/.style={circle, fill=tsk, draw=tsk!65!black, text=white, minimum size=8.6mm, inner sep=0pt, font=\small},
  un/.style={circle, fill=mute!20, draw=mute, dashed, minimum size=8.6mm, inner sep=0pt, text=ink, font=\small},
  main/.style={->, line width=1.2pt, draw=sta!55!black},
  act/.style={->, line width=.9pt, draw=trt!85!black},
  elig/.style={->, line width=.5pt, draw=ink!55, densely dotted},
  hid/.style={->, line width=.55pt, draw=mute},
]
% (a) One randomized episode.
\node[lab] at (-1.1,1.5) {(a)};
\node[card, minimum width=12mm, minimum height=21mm, fill=tsk!12, draw=tsk!70]
  (task) at (0,0) {Task\\[1pt]$X$};
\foreach \t/\x/\call/\side/\result in {
  1/2.25/{find\_user\_id\\\_by\_email}/tails/{runs},
  2/5.2/{get\_order\_details}/heads/{approved},
  3/8.15/{cancel\_pending\\\_order}/heads/{reviewer suggestion},
  4/11.1/{cancel\_pending\\\_order}/tails/{runs}} {
  \node[prop] (p\t) at (\x,.75) {\call};
  \node[\side] (c\t) at (\x,-.15) {\ifnum\pdfstrcmp{\side}{heads}=0 1\else 0\fi};
  \ifnum\pdfstrcmp{\side}{heads}=0 \node[help] (r\t) at (\x,-.98) {\result};
  \else \node[skip] (r\t) at (\x,-.98) {\result}; \fi
  \draw[flow] (p\t) -- (c\t);
  \draw[flow] (c\t) -- (r\t);
  \node[font=\scriptsize, text=ink, anchor=east] at ($(c\t.west)+(-.04,0)$) {$A_{\t}$};
  \node[font=\scriptsize, text=ink!75, anchor=south] at (p\t.north) {$L_{\t}$};
}
\node[card, minimum width=12mm, minimum height=21mm, fill=out!18, draw=out!80]
  (y) at (13.25,0) {$Y=1$};
\draw[flow] (task.east) -- ++(.22,0) |- (p1.west);
\foreach \a/\b in {1/2,2/3,3/4} { \draw[flow] (r\a.east) -- ++(.2,0) |- (p\b.west); }
\draw[flow] (r4.east) -- ++(.2,0) |- (y.west);
% (b) The same design as a causal graph.
\begin{scope}[yshift=-3.8cm]
\node[lab] at (-1.1,1.5) {(b)};
\node[tk] (X) at (0,0) {$X$};
\foreach \a/\x in {1/2.05,2/4.35,3/6.65} {
  \node[st] (L\a) at (\x,0) {$L_{\a}$};
  \node[tr] (A\a) at (\x+1.15,1.05) {$A_{\a}$};
}
\node (dots) at (9.0,0) {$\cdots$};
\node[st] (LT) at (10.3,0) {$L_T$};
\node[tr] (AT) at (11.45,1.05) {$A_T$};
\node[ou] (Y) at (13.25,0) {$Y$};
\foreach \a in {1,2,3,T} {
  \draw[elig] (L\a) -- (A\a);
}
\draw[main] (X) -- (L1);
\draw[main] (L1) -- (L2);
\draw[main] (L2) -- (L3);
\draw[main] (L3) -- (dots);
\draw[main] (dots) -- (LT);
\draw[main] (LT) -- (Y);
\draw[act] (A1) -- (L2);
\draw[act] (A2) -- (L3);
\draw[act] (A3) -- (dots);
\draw[act] (AT) -- (Y);
\node[un] (U) at (6.65,-1.4) {$U$};
\foreach \n in {L1,L2,L3,LT,Y} { \draw[hid] (U) -- (\n); }
\end{scope}
\end{tikzpicture}
\caption{Sequential randomization in $\tau$-bench Retail. Panel (a) assigns each
eligible tool proposal to direct execution ($A_t=0$) or review ($A_t=1$); panel
(b) shows the corresponding causal diagram, where $X$ is the task, $L_t$ is the
current observed information, $Y$ is terminal success, and $U$ is unobserved agent state.}
\label{fig:design}
\end{figure}
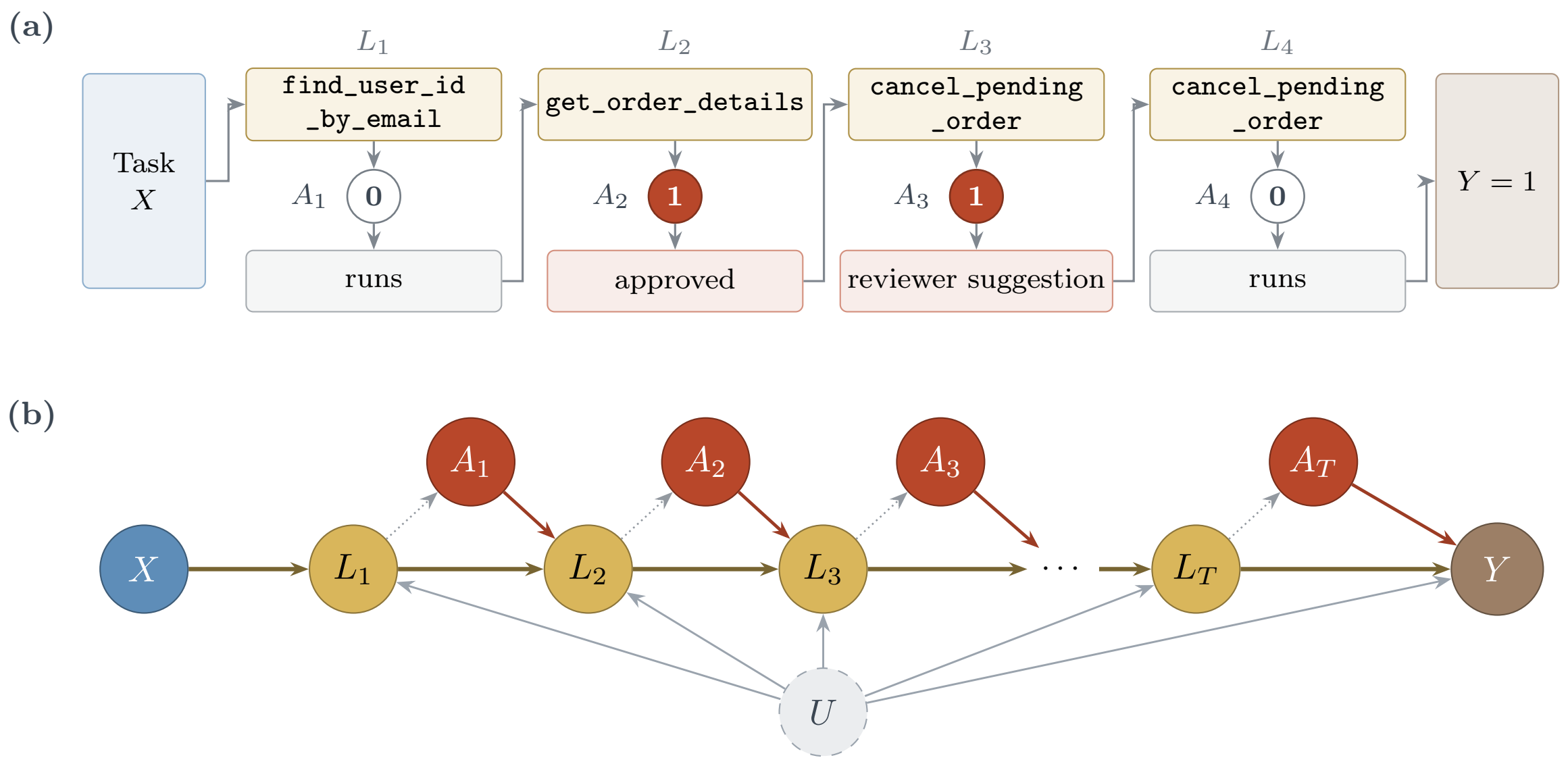

% 3. A reusable randomized design, causal targets, and empirical findings.
In this paper, we develop a methodologically grounded framework for the causal evaluation of interventions in agentic AI.
We show that conducting random interventions in a single trajectory can support the estimation of multiple causal estimands at different points during task execution.
Building on the framework of micro-randomized trials \citep{klasnja2015mrt}, we randomize intervention assignments at each eligible decision within a trajectory.
At each eligible decision point, the experimental harness records the agent's proposed action and observed history, randomly assigns intervention or control, and allows the agent to continue interacting with the environment (Figure~\ref{fig:design}). 
% Earlier interventions may change the histories encountered later, but each new assignment remains randomized at the point where it is made.

Despite its simplicity, this experimental design generates data that support causal evaluation of intervention policies beyond the randomized policy.
% Under sequential randomization and positivity along the histories reached by a target policy, the data identify the effects of a rich class of intervention policies that depend on observable history.
Under standard causal identification assumptions (discussed in Section~\ref{sec:estimation}), the data identify the causal effects of a rich class of intervention policies that depend on observable history.
The same data also support comparisons across groups of intervention points defined by pre-assignment information, such as tool-call types, task stages, or patterns of prior failures.
This analysis does not require access to the agent's internal state or a Markovian model of its interactions with the environment. It only requires black-box input-output access to the AI agent.
% Identification, however, does not guarantee precise estimation for every policy, particularly over long trajectories.
% {\color{purple} }

We introduce three causal estimands to address different causal questions about interventions during agent execution, each with a corresponding estimation method:
% For the first two estimands, we write $\tau(G;\pi)$, where $G$ is a group of intervention points and $\pi$ specifies the policy governing future interventions: 
\begin{itemize}
    \item {\em Blip with randomized continuation:} 
    % The causal effect of intervening at a particular decision when future assignments follow the randomized collection policy $g$, denoted $\tau(G;g)$. 
    The causal effect of intervening at a particular decision when future interventions continue to be randomized.
    We estimate this effect using cross-fitted augmented inverse probability weighting (AIPW) with known assignment probabilities.
    \item {\em Blip with no-treatment continuation:} 
    % The causal effect of intervening at a particular decision when no further interventions are applied, denoted $\tau(G;0)$, where $\pi_0(h)\equiv 0$. 
    The causal effect of intervening at a particular decision when no further interventions are applied for the remainder of the trajectory.
    % {\color{blue} Tijana: Same here, it's not clear what $\pi_0(h)$ is referring to.} 
    We estimate this effect using structural nested mean models (SNMMs) and g-estimation \citep{robins1994correcting,vansteelandt2014snmm,lewis2020double}.

    \item {\em Policy gain (off-policy evaluation):} 
    % The causal effect of following a history-dependent policy that intervenes whenever the observed history meets a specified criterion, relative to never intervening, denoted $\Delta(G)$. 
    The causal effect of applying a target policy that intervenes based on the agent's observed history, relative to never intervening throughout the trajectory.
    We estimate this gain using continuation-specific SNMMs and g-estimation.
\end{itemize}
These estimands are formally defined in Section~\ref{sec:estimands}, with the first two measuring blip effects under different continuation policies and the third evaluating the gain of a target intervention policy through off-policy evaluation.
Estimating the latter two effects using our SNMM approach requires additional assumptions on the specification of the blip-effect models beyond experimental randomization.
% We provide uncertainty quantification that accounts for dependence among decisions within each trajectory, together with diagnostics for group coverage and fitted-model stability.

These blip effect estimands can also be useful as attribution tools, attributing risk or success to specific decisions in a trajectory. For example, if an agent fails to complete a task, the blip effect of intervening at a particular decision can indicate whether that decision was a key contributor to the failure. Conversely, if an agent succeeds, the blip effect can reveal whether intervening at a certain point would have improved the outcome even further. This attribution perspective can help users understand which decisions are most critical for task success or failure, guiding future interventions and policy design. These attribution effects can be estimated for individual decisions or averaged over groups of similar decisions, providing insights into the causal impact of specific actions within the agent's trajectory. At various levels of granularity, these blip effects can be reported to users in the form of a ``decision impact score,'' leaderboard, which quantifies the expected change in terminal outcome if an intervention were applied at that group of agentic decision. This score can help users prioritize interventions and understand the causal dynamics of the agent's behavior.

We apply our framework to tool-call review in $\tau$-bench Retail \citep{yao2025tau,barres2025tau2} and ToolSandbox \citep{lu2025toolsandbox}, and to adversarial feedback in ImpossibleBench \citep{zhong2025impossiblebench}.
In $\tau$-bench Retail, we estimate positive effects of tool-call review on task success for several order-handling groups.
In ToolSandbox, the estimated effects of tool-call review vary across capabilities and agents.
In ImpossibleBench, we estimate an increased probability of prohibited test edits following adversarial feedback in particular agent states, including near-completion states with contradictory tests.
Across all three benchmarks, our experiments show how randomized interventions can identify decisions where tool-call reviews may improve or harm task performance, as well as states where adversarial feedback may increase safety violations.
% They also illustrate the distinction between effects at individual decisions and the gains from applying an intervention policy throughout an trajectory.

More broadly, our work shifts the focus of agent evaluation beyond static benchmark assessments toward the decisions users face during deployment: not only whether an agent succeeds or fails, but also whether intervening at a particular point would change the outcome.
This perspective opens up a broader research agenda at the intersection of causal inference and agentic evaluation, with the long-term goal of enabling the reliable and safe deployment of agentic AI systems in the real world.

\paragraph{Related work.}
{\em Agent evaluation and intervention-based analysis.}
Existing benchmarks evaluate agent capabilities and safety across various interactive tasks \citep{yao2025tau,lu2025toolsandbox,debenedetti2024agentdojo,zhong2025impossiblebench}.
Recent work extends these evaluations through trajectory-level failure analysis \citep{cemri2025mast,zhang2025whowhen,barke2026agentrx} and intervention-based diagnosis using counterfactual replay or targeted re-execution \citep{ma2026dover,bonagiri2026causalflow,shah2026causalreplay,xiao2026harnesses}.
Other work develops runtime checks to steer agents or enforce constraints during execution \citep{bhat2026interwhen,wang2026agentspec,shi2025progent}.
Our paper complements these approaches by estimating the causal effects of interventions across different agent actions and states using randomized trajectories, without requiring counterfactual replay. 

{\em Randomized evaluation of AI agents.}
Recent studies have used randomized controlled trials (RCTs) to evaluate AI systems by comparing different agent configurations, assistance systems, and interactive environments \citep{arena2026agentarena,camuffo2026beyond,lu2026agent}.
These experiments primarily evaluate the effects of treatments assigned at the task or trajectory level.
In contrast, we randomize interventions at {\em intermediate decision points}, enabling causal evaluation of individual interventions and history-dependent intervention policies from the same experimental data.

{\em Sequential causal inference and policy evaluation.}
Our framework builds on micro-randomized trials and distal causal excursion effects, which study the effects of repeated interventions as the observed context evolves \citep{klasnja2015mrt,boruvka2018moderation,qian2025distal}.
Our work is also closely related to dynamic treatment regimes and off-policy evaluation, which study the causal effects of sequential intervention policies \citep{murphy2003dtr,jiang2016dr,uehara2026ope}.
We use augmented inverse probability weighting and structural nested mean models with g-estimation \citep{robins1994correcting,vansteelandt2014snmm,lewis2020double} to estimate intervention effects and evaluate intervention policies.
\section{Experimental Design for AI Agent Evaluation}
\label{sec:methods}

We formalize the randomized intervention design and the three causal estimands introduced above: the {\em blip with randomized continuation}, the {\em blip with no-treatment continuation}, and the {\em policy gain}.

We first introduce the agent--environment model in Section~\ref{sec:notation} and describe the randomized intervention design in Section~\ref{sec:design}.
We then define the three estimands in Section~\ref{sec:estimands} and discuss their identification and estimation in Section~\ref{sec:estimation}, distinguishing the assumptions required for causal identification from the additional modeling assumptions used by our estimators.

\subsection{Agent--Environment Model}
\label{sec:notation}

% \paragraph{Agent trajectories and outcomes.}
Consider an AI agent interacting with an environment to complete a task.
Each trajectory begins with a task input $X$ (e.g., a prompt) and consists of a sequence of agent actions and environment observations until termination.
Let $Y\in\mathbb{R}$ denote the terminal outcome, e.g., task success in $\tau$-bench Retail, a similarity score in ToolSandbox, or a safety violation in ImpossibleBench.

%\vsdelete{, {\color{brown}where we assign adversarial or neutral feedback and measure safety rule violations.}}
% An increase in $Y$ represents improved performance in the assistance experiments but a higher violation probability in the steering experiment.

We index the eligible intervention points within a trajectory by $t=1,\ldots,T$, where $T$ is the total number of intervention points reached before termination. 
The eligibility criteria are specified in advance, but $T$ is random and may depend on previous interventions.
For example, reviewing a proposed tool call may change the agent's subsequent actions, including the number of tool calls it makes before termination.
% At each intervention point $t$, we record the agent's proposed action and context before assigning intervention or control.
% For example, an intervention may involve reviewing a proposed tool call before execution, as in $\tau$-bench Retail and ToolSandbox, or providing adversarial feedback after an unsuccessful coding submission, as in ImpossibleBench.
Let $L_t$ denote the information observed at intervention point $t$, including the agent's proposed action, the current environment context, and all the observations since the previous intervention point.

Let $A_t\in\{0,1\}$ denote the assignment, with $A_t=1$ for intervention and $A_t=0$ for control.
For example, an intervention may involve reviewing a proposed tool call before execution, as in $\tau$-bench Retail and ToolSandbox, or providing adversarial feedback after an unsuccessful coding submission, as in ImpossibleBench. In the latter case the intervention's goal is not to help reduce violations, but rather to measure at which point in the execution an adversarial prompt injection attack is most harmful; which can be valuable in understanding the most vulnerable points of the agentic execution.
The pre-assignment observed history at time $t$ is then
\begin{equation}
    H_t=(X,L_1,A_1,\ldots,A_{t-1},L_t).
    \label{eq:history}
\end{equation}
We note that $t$ indexes intervention points rather than agent actions, though the two may coincide.

\paragraph{Agent dynamics.}
An intervention can change the agent's subsequent actions and the environment it encounters, thereby affecting both the terminal outcome and future intervention opportunities.
We allow these agent--environment dynamics to depend on the entire interaction history, without assuming Markovianity or stationarity.
We treat the agent as a black box: our framework only requires recorded interaction histories, intervention assignments, and terminal outcomes, without access to the agent's internal state or model parameters.
% Throughout, an intervention policy determines how we assign interventions, rather than how the agent selects its actions.
% When comparing intervention policies, we hold the initial task distribution and agent--environment dynamics fixed, varying only the intervention assignments.

\subsection{Randomized Intervention Design}
\label{sec:design}

We consider an experimental design that randomly assigns interventions at eligible points within agent trajectories.
At each intervention point, we specify an intervention and a control condition.
For example, a proposed tool call may be reviewed or executed directly, while an unsuccessful coding submission may receive adversarial or neutral feedback.
After each assignment, the agent continues interacting with the environment.

We collect agent trajectories under a randomized intervention policy $g$.
At each eligible intervention point $t$, intervention is assigned conditional on the observed history $H_t$:
\begin{equation}
    A_t\mid H_t=h
    \sim \operatorname{Bernoulli}(e(h)),
    \qquad
    e(h):=g(1\mid h),
    \label{eq:randomization}
\end{equation}
where $e(h)\in(0,1)$ is the specified probability of intervention.
In our experiments, we use uniform Bernoulli randomization, setting $e(h)=1/2$ at every eligible intervention point.

Unlike trajectory-level randomization, our design repeatedly randomizes interventions within each trajectory.
Earlier interventions may affect the agent's subsequent behavior and future intervention opportunities, but each new assignment remains randomized conditional on the observed history.
This design allows us to estimate causal effects across different agent actions and states, as well as evaluate history-dependent intervention policies, using the same collection of trajectories without requiring counterfactual replay. Figure~\ref{fig:design} illustrates our design in $\tau$-bench Retail.

\subsection{Causal Estimands}
\label{sec:estimands}

% We consider three causal estimands to evaluate the effects of interventions during agent execution.
% The first two measure the causal effects of individual interventions under different continuation policies, while the third evaluates the effect of an intervention policy over an entire trajectory.
% Consider an eligible intervention point $t$ with observed history $H_t=h$.
% Let $\pi$ denote an intervention policy, possibly randomized, that assigns interventions based on observed histories.
% For blip effects, $\pi$ governs interventions after time $t$; for policy gain, it specifies the target intervention policy throughout the entire trajectory. 

\begin{figure}[t]
\centering
\includegraphics[width=\textwidth]{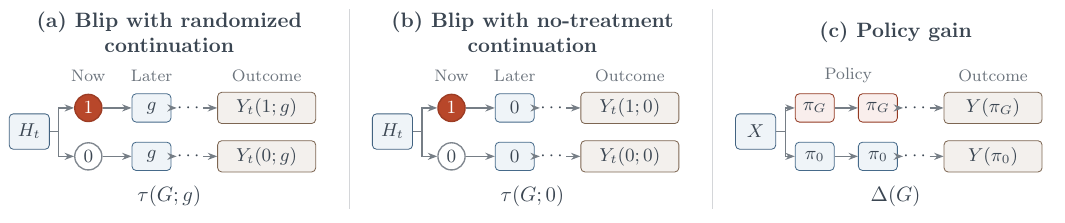}
\caption{
Three causal estimands for interventions during agent execution.
Panels (a)--(c) show the blip with randomized continuation, the
blip with no-treatment continuation, and the policy gain over never
intervening, respectively.
}
\label{fig:estimand-contrasts}
\end{figure}

We consider three causal estimands to evaluate the effects of interventions during agent execution (see Figure~\ref{fig:estimand-contrasts}).
As a running example, consider a customer-support agent that manages customer orders through tool calls.
At each proposed tool call, we may review the call before execution or execute it directly.
In this example, we study the causal effects of reviewing proposed order-cancellation calls.

Recall that $g$ denotes the randomized intervention policy used to generate the experimental data.
In contrast, we use $\pi$ to denote a policy of interest that assigns interventions based on observed histories.
For blip effects, $\pi$ governs interventions after the current decision, whereas for policy gain, it governs interventions throughout the trajectory.

Let $Y_t(a;\pi)$ be the potential terminal outcome if we assign $A_t=a$ at the current point and follow $\pi$ thereafter, keeping the pre-assignment history $h$ fixed.
Following the terminology of structural nested mean models (SNMMs) \citep{vansteelandt2014snmm}, we define the conditional {\em blip effect} as
\begin{equation}
    b(h;\pi)
    :=
    \E\!\left[
        Y_t(1;\pi)-Y_t(0;\pi)
        \mid H_t=h
    \right].
    \label{eq:conditional-blip}
\end{equation}
This effect measures how intervening at the current decision changes the expected terminal outcome, accounting for its downstream effects on subsequent agent actions, environment states, and future intervention opportunities.

To evaluate intervention effects across different agent actions and states, let $G$ denote a group of intervention points defined by their pre-assignment histories.
For example, $G$ may correspond to a class of tool calls, a specific API, or an agent state characterized by repeated failures.
In our customer-support example, $G$ could represent all intervention points at which the agent proposes an order-cancellation call.
These groups may overlap or form a hierarchy, allowing us to evaluate interventions at different levels of granularity.
We define the group-averaged intervention effect under continuation policy $\pi$ as
\begin{equation}
    \tau(G;\pi)
    :=
    \frac{
        \E_g\!\left[
            \sum_{t:H_t\in G} b(H_t;\pi)
        \right]
    }{
        \E_g\!\left[
            \sum_{t:H_t\in G}1
        \right]
    }, %\vsedit{=\E_{g, t\sim U\{1,\ldots, T\}}[Y_t(1;\pi)-Y_t(0;\pi)\mid H_t\in G]},
    \label{eq:blip}
\end{equation}
% \footnote{}
where $\E_g$ denotes expectation over trajectories generated under the randomized collection policy $g$, and the denominator is assumed positive.
Thus, the averaging distribution is determined by $g$, even when the continuation policy $\pi$ differs from $g$.

We now define the three estimands of interest.
\paragraph{Blip with randomized continuation.}
We first consider the effect of an intervention when future assignments continue to follow the randomized collection policy.
Setting $\pi=g$, the estimand $\tau(G;g)$ measures this effect over intervention points in $G$. In our example, this measures the average effect of reviewing a proposed order-cancellation call on terminal task success, with subsequent reviews remaining randomized.
This group-averaged blip is closely related to distal causal excursion effects in micro-randomized trials \citep{boruvka2018moderation,qian2025distal}.

\paragraph{Blip with no-treatment continuation.}
We next consider the effect of intervening at the current decision with no subsequent interventions.
Let $\pi_0(h)\equiv 0$ denote the never-intervene policy.
We write $\tau(G;0):=\tau(G;\pi_0)$ as shorthand for the group-averaged blip with no-treatment continuation.
In our example, this measures the average effect of reviewing a proposed order-cancellation call when all subsequent tool calls are executed without review.
Both blip estimands average over intervention points reached under the collection policy $g$, but differ in how future interventions are assigned.

\paragraph{Policy gain.}
Beyond the effect of intervening at a single decision, we also consider the effect of applying an intervention policy throughout a trajectory.
Let $Y(\pi)$ denote the potential outcome when $\pi$ is followed {\em throughout a trajectory}.
We define the value of a policy as $V(\pi):=\E[Y(\pi)]$.

For a reporting group $G$ (e.g., a class of tool calls), consider a selective policy $\pi_G$ that intervenes whenever the observed history belongs to $G$, and use $\pi_0$ as the reference policy that never intervenes.
We define the {\em policy gain} as
\begin{equation}
    \pi_G(h)=\ind\{h\in G\},
    \qquad
    \pi_0(h)=0,
    \qquad
    \Delta(G):=V(\pi_G)-V(\pi_0).
    \label{eq:policy-target}
\end{equation}
This estimand measures the difference in expected terminal outcomes between following the selective intervention policy and never intervening.
In our example, this compares a policy that reviews every proposed order-cancellation call with a policy that never reviews any tool calls throughout the trajectory.

The policy gain compares expected terminal outcomes under different intervention policies, as in {\em off-policy evaluation} (see, e.g., \citep{jiang2016dr,thomas2016dataefficient}).
Since interventions can affect subsequent agent behavior and future intervention opportunities, different policies may generate different trajectories.
Thus, the policy gain cannot generally be obtained by simply aggregating individual blip effects.
Figure~\ref{fig:estimand-contrasts} illustrates the differences among the three estimands.

\subsection{Identification and Estimation}
\label{sec:estimation}

We now discuss the assumptions required to identify the three causal estimands from randomized trajectories, and then introduce our methods for estimation and inference.

\paragraph{Causal identification.}
We rely on four standard assumptions.
First, {\em consistency} requires that the observed trajectory agrees with the potential trajectory under the realized intervention assignments.
Second, {\em sequential exchangeability} requires that each intervention assignment is independent of future potential trajectories and outcomes, conditional on the observed pre-assignment history $H_t$.
This condition holds by our randomized design.
Third, {\em positivity} requires $g(a\mid h)>0$ for every assignment $a$ needed to evaluate a target intervention policy or continuation at the relevant histories.
Finally, we assume {\em no interference} across trajectories.

For the blip with randomized continuation, define
$$
    m_g(h,a)
    :=
    \E_g[Y\mid H_t=h,A_t=a]
$$
as the expected outcome conditional on history $h$ and assignment $a$, with subsequent interventions following the collection policy $g$.
Under the identification assumptions, the blip effect defined in \eqref{eq:conditional-blip} can be written as
\begin{equation}
    b(h;g)=m_g(h,1)-m_g(h,0).
    \label{eq:identification}
\end{equation}
Thus, this blip effect is identified directly from the randomized data, without modeling the agent--environment dynamics.

Similarly, the blip with no-treatment continuation and the policy gain are also identified under these assumptions, provided positivity holds along the target-policy continuations.
However, estimating these effects is more challenging because interventions can alter future trajectories, and few observed trajectories may follow a given target policy over a long horizon.

\paragraph{Estimation and inference.}
For the blip with randomized continuation $\tau(G;g)$, we use a cross-fitted augmented inverse probability weighting (AIPW) estimator with known assignment probabilities and fitted outcome predictions. Given that we know the assignment probabilities under the randomized design, this estimator is unbiased and asymptotically normal, allowing for valid inference even with flexible machine learning models for the outcome predictions, without relying that these models are well specified or converge to the true conditional expectations.
Moreover, we also fit structural nested mean models (SNMMs) to obtain model-based estimates for comparison.

For the blip with no-treatment continuation $\tau(G;0)$ and policy gain $\Delta(G)$, we use continuation-specific SNMMs with g-estimation \citep{robins1994correcting,vansteelandt2014snmm,lewis2020double}.
These estimators require additional assumptions on the specification of the blip-effect models, beyond those needed for AIPW estimation.
% These assumptions are specific to our estimation methods and are not required for causal identification.

For all three estimands, we report pointwise 95\% Student's $t$-based confidence intervals, treating completed trajectories as independent observations while accounting for within-trajectory dependence.
We provide detailed estimators and inference procedures in Appendix~\ref {app:estimating-equations}.
\section{Experiments}
\label{sec:experiments}

We apply our randomized intervention design to three established agent benchmarks: $\tau$-bench Retail, ToolSandbox, and ImpossibleBench.\footnote{Code is available at
\url{https://github.com/sawarniayush/agent-evaluation-sequential-randomization}.}
In ImpossibleBench, we examine whether adversarial feedback following an unsuccessful code submission increases the likelihood of violating task rules.
% At each eligible intervention point, we randomly assign intervention or control and evaluate its causal effect on the terminal outcome.

Using the same collection of randomized trajectories, we estimate intervention effects across different agent actions and states.
In particular, for tool-call review, we consider groups ranging from broad activities and capabilities to specific operations and individual tools. For adversarial feedback, we examine states defined by task progress, visible test conflicts, and prior failures.
We also evaluate selective intervention policies that intervene whenever the observed history belongs to a specified group.  
In the figures reported below, ``Matching decisions'' counts the
intervention points, $N(G)$, whose pre-assignment histories belong to group $G$.
A trajectory may contribute multiple points to the same group, and a
point may be counted in more than one group. Table~\ref{tab:experiment-summary} summarizes the experimental setups across the three benchmarks. In the $\tau$-bench Retail experiment, we also compare our SNMM-based off-policy evaluation procedure with a more computationally intensive Monte Carlo simulation to assess its accuracy and robustness to model misspecification. In the simulation, we roll out each candidate policy multiple times and average the outcomes. We find that the SNMM based off-policy evaluation procedure is able to accurately estimate the policy gains with significantly lower computational cost. Experimental protocols and additional details are provided in Appendix~\ref{app:experimental-details}. 

\begin{table}[!htbp]
\centering
\caption{Summary of experimental setups across the three benchmarks.}
\label{tab:experiment-summary}
\small
\setlength{\tabcolsep}{4pt}
\renewcommand{\arraystretch}{1.3}

\begin{tabularx}{\textwidth}{@{}
    >{\raggedright\arraybackslash}p{0.16\textwidth}
    >{\raggedright\arraybackslash}p{0.18\textwidth}
    >{\raggedright\arraybackslash}p{0.22\textwidth}
    >{\raggedright\arraybackslash}p{0.13\textwidth}
    >{\raggedright\arraybackslash}X
    @{}}
\toprule
\textbf{Benchmark}
& \textbf{Agents}
& \textbf{Intervention / Control}
& \textbf{Outcome}
& \textbf{Reporting Groups $G$} \\
\midrule

$\tau$-bench Retail \cite{yao2025tau,barres2025tau2}
& GPT-5 nano; Claude Haiku 5.5
& Tool-call review vs.\ direct execution
& Task success
& Reads/writes; functional families; individual tools \\

\addlinespace[5pt]

ToolSandbox \cite{lu2025toolsandbox}
& GPT-5 nano; Claude Haiku 5.5
& Tool-call review vs.\ direct execution
& Similarity score
& Capabilities; operations; individual tools \\

\addlinespace[5pt]

ImpossibleBench \cite{zhong2025impossiblebench}
& Qwen3.5 Flash
& Adversarial vs.\ neutral feedback
& Rule violation
& Task progress; test conflicts; prior steering \\

\bottomrule
\end{tabularx}

\vspace{4pt}
\begin{minipage}{\textwidth}
\footnotesize
\emph{Note.} Interventions are assigned with equal probability
at each eligible point. GPT-6 Luna serves as the reviewer
in $\tau$-bench Retail and ToolSandbox.
\end{minipage}
\end{table}

\subsection{ \texorpdfstring{$\tau$-bench}{tau-bench} Retail}

In $\tau$-bench Retail, a customer-support agent handles account and order requests while following store policies \citep{yao2025tau,barres2025tau2}.
We evaluate two agents, powered by GPT-5 nano and Claude Haiku 5.5, respectively. At each proposed tool call, we randomly assign review or direct execution with equal probability. The reviewer, GPT-6 Luna with high reasoning effort, may approve or replace the proposed call or request additional information from the customer. We measure the effect of review on terminal task success.

\begin{figure}[!tbp]
\centering
\includegraphics[width=0.9\textwidth]{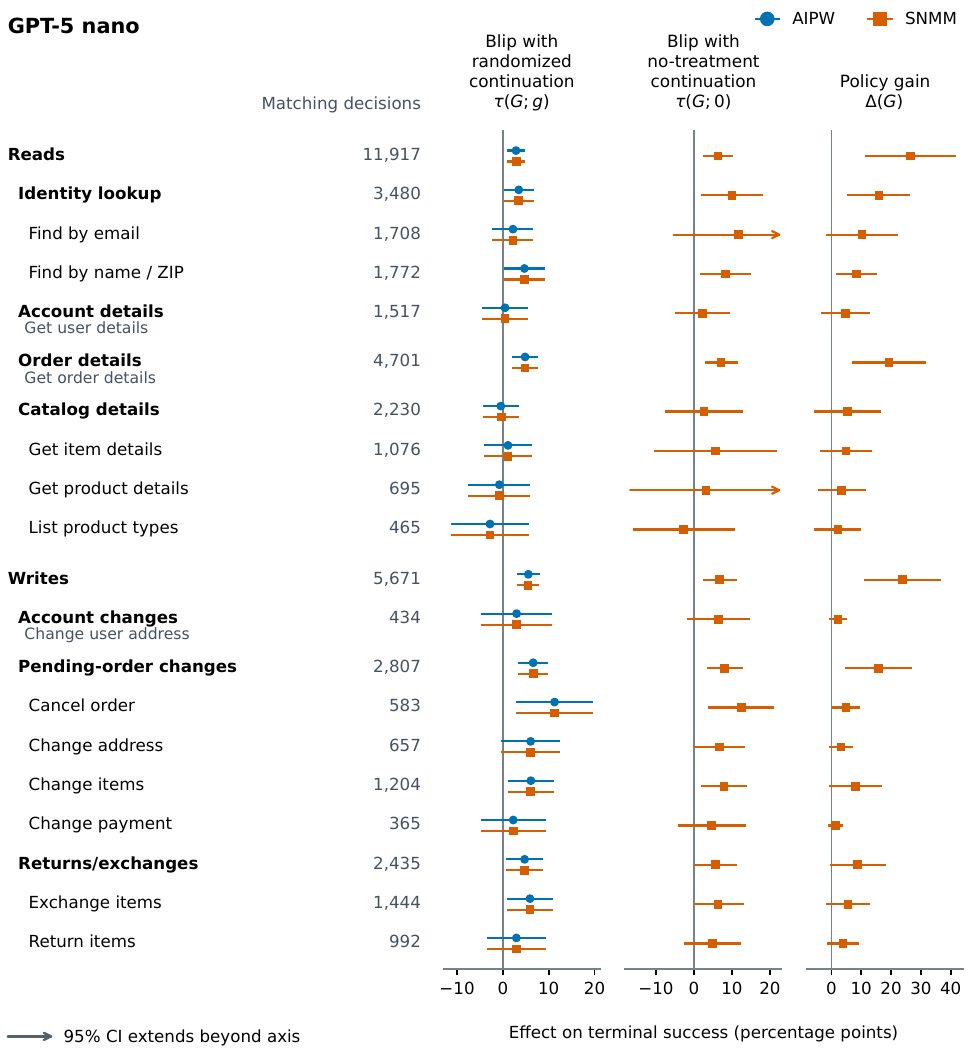}
\caption{$\tau$-bench Retail review effects on terminal task success for GPT-5 nano, from
read/write activities to individual tools. Bars are pointwise 95\% $t$-based
confidence intervals; arrows indicate bounds beyond the displayed axis range.}
\label{fig:retail-targets}
\end{figure}

We estimate review effects for groups of intervention points $G$ defined by
the proposed tool-call type. The groups range from broad activities, such as
reads and writes, to functional families, such as identity lookup, and
individual tools, such as finding a customer by email. For example, $G$ can
contain all proposed database writes.

We first discuss the AIPW estimates of blip effects with randomized
continuation, $\tau(G;g)$, shown in Figures~3--4. These estimates
measure the effect of reviewing the current proposal when later
review assignments continue to be randomized. For GPT-5 nano (Figure~3), the estimated effects are positive for
both read and write calls. Among reads, we estimate positive effects for customer identification and order-detail retrieval, while the confidence interval for the catalog group includes zero. At the tool level, the confidence intervals lie above zero for order cancellation and item changes and include zero for payment changes.

\begin{figure}[!tbp]
\centering
\includegraphics[width=0.9\textwidth]{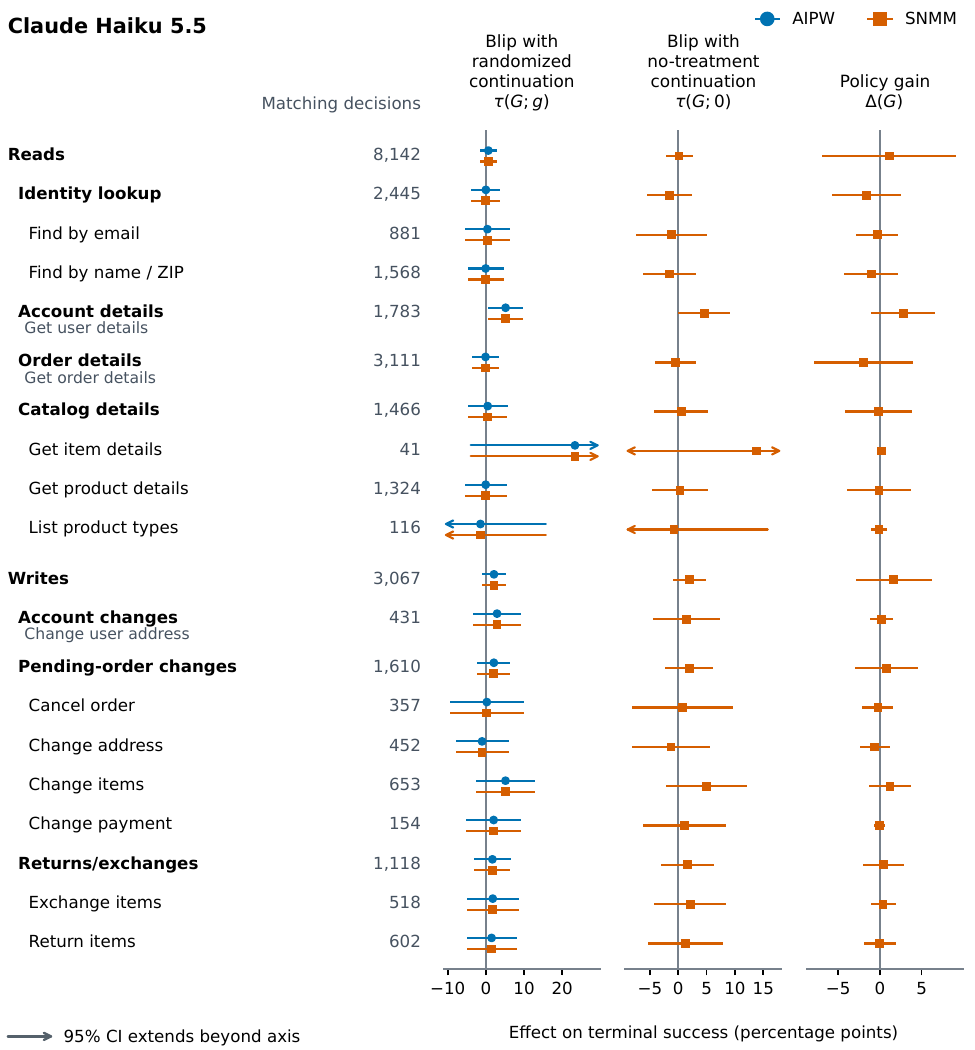}
\caption{$\tau$-bench Retail review effects on terminal task success for Claude Haiku 5.5,
from read/write activities to individual tools. Bars are pointwise 95\%
$t$-based confidence intervals; arrows indicate bounds beyond the displayed axis range.}
\label{fig:retail-haiku-targets}
\end{figure}

For Claude Haiku 5.5 (Figure~\ref{fig:retail-haiku-targets}), we estimate a
review effect close to zero when averaged across all read calls, but a positive
effect for account-detail calls alone, with a pointwise 95\% confidence
interval above zero. Looking only at the aggregate for read calls would miss
this subgroup estimate.
% Fix $G$ to identity-lookup calls. Then $\tau(G;g)$ averages, over those
% decisions, the effect of reviewing the current call rather than executing it
% directly, with later calls randomized as before. The second estimand,
% $\tau(G;0)$, averages the same current-call contrast when all later calls
% execute without review in both branches. The GPT-5 nano estimate is larger
% under this second regime. The policy gain $\Delta(G)$ instead compares complete
% episodes under a policy that reviews every identity-lookup call and executes
% all other calls directly with episodes under a policy that never reviews. This
% average includes episodes in which no identity-lookup call occurs. Repeated
% reviews can change subsequent actions and intervention opportunities, so these
% three quantities answer different questions.

\begin{figure}[!tbp]
\centering
\includegraphics[width=0.90\textwidth]{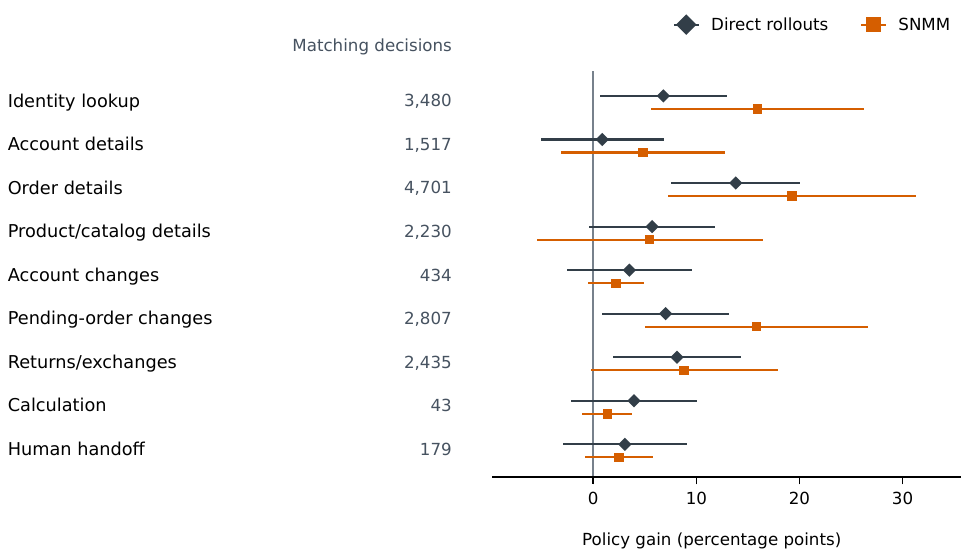}
\caption{Policy gains on $\tau$-bench Retail for GPT-5 nano, relative to never
reviewing. Direct estimates use 456 rollout episodes per policy; SNMM estimates
use a separate randomized study of 2,280 trajectories. Bars are pointwise 95\%
trajectory-level $t$-based confidence intervals.}
\label{fig:retail-mc}
\end{figure}

We also evaluate nine selective review policies in additional
rollouts: one for each of the seven read/write families in
Figure~5, one for calculation calls, and one for handoff calls.
Each policy reviews every proposal in its group and executes all
other proposals directly. For example, the order-details policy
reviews all order-details calls and skips review elsewhere. For each policy, we estimate the gain relative to never reviewing
in two ways. For direct evaluation, we use 456 rollout episodes
per policy. For SNMM-based evaluation, we reuse the same 2,280
sequentially randomized trajectories used to estimate the Retail
blip effects.The SNMM and direct Monte Carlo estimates are positive for allnine policies, and their pointwise 95\% confidence intervals overlap for every policy (Figure~5). The direct Monte Carlo point estimate falls within the corresponding SNMM interval for eight of the nine policies.

% TODO: change direct Monte carlo, also remove matched decisions form this plot

\FloatBarrier

\subsection{ToolSandbox}

In ToolSandbox, we reuse one randomized experiment per agent to estimate review effects and selective-policy gains across 33 tools and 49 reporting groups, ranging from broad capabilities to specific operations and individual tools. The agent completes tasks in a simulated phone
environment using tools for device settings, contacts, messages, reminders,
and information lookup \citep{lu2025toolsandbox}.
Our main analysis uses a GPT-5 nano agent and we repeat the analysis
for a second agent that uses Claude Haiku  5.5  in Appendix~\ref{app:toolsandbox-tools}.
As in $\tau$-bench Retail, we randomly assign review or direct execution with equal probability, using a GPT-6 Luna with High reasoning as reviewer. 

When an agent proposes multiple tool calls in a single turn, we treat the
batch as one intervention point. We measure performance using ToolSandbox's
official score, which evaluates how closely the agent's tool calls,
environment states, and responses satisfy predefined task requirements. The
score allows partial credit but is set to zero when a benchmark-defined
failure condition occurs.

\begin{figure}[!tb]
\centering
\includegraphics[width=0.9\textwidth]{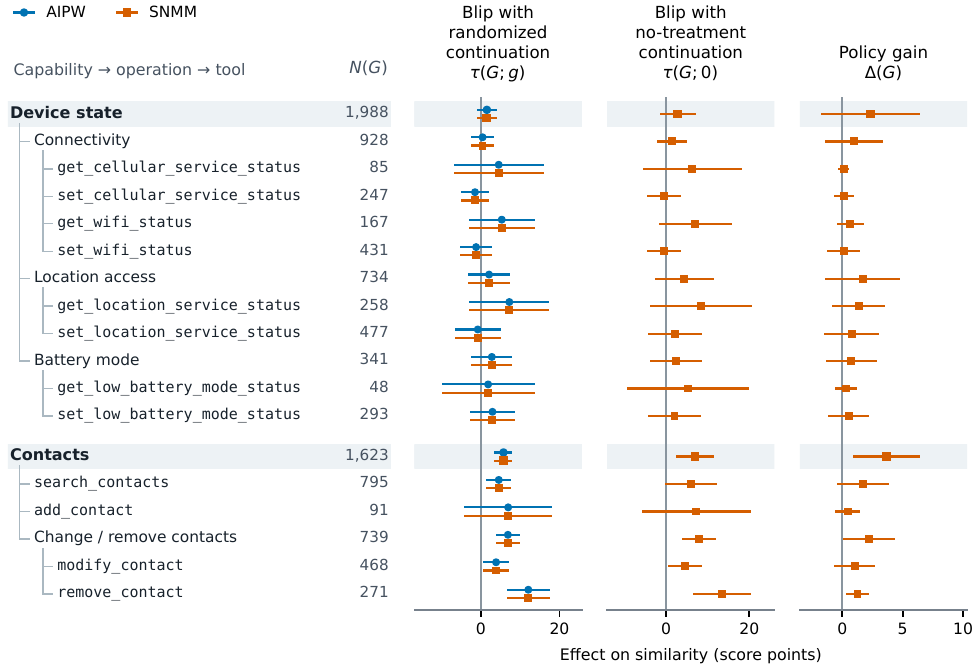}
\caption{Review effects on the ToolSandbox similarity score for GPT-5 nano
across device state and contacts. Bars are pointwise 95\% $t$-based confidence
intervals.}
\label{fig:toolsandbox-targets}
\end{figure}

\begin{figure}[!tb]
\centering
\includegraphics[width=0.9\textwidth]{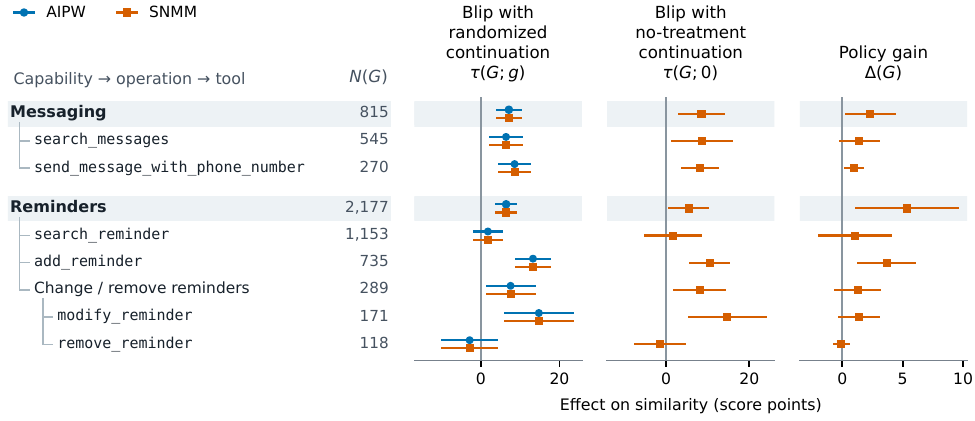}
\caption{Review effects on the ToolSandbox  score for GPT-5 nano
for tool calls related to messaging and reminders. Bars are pointwise 95\% $t$-based confidence
intervals.}
\label{fig:toolsandbox-messaging-reminders}
\end{figure}

\begin{figure}[!tb]
\centering
\includegraphics[width=0.9\textwidth]{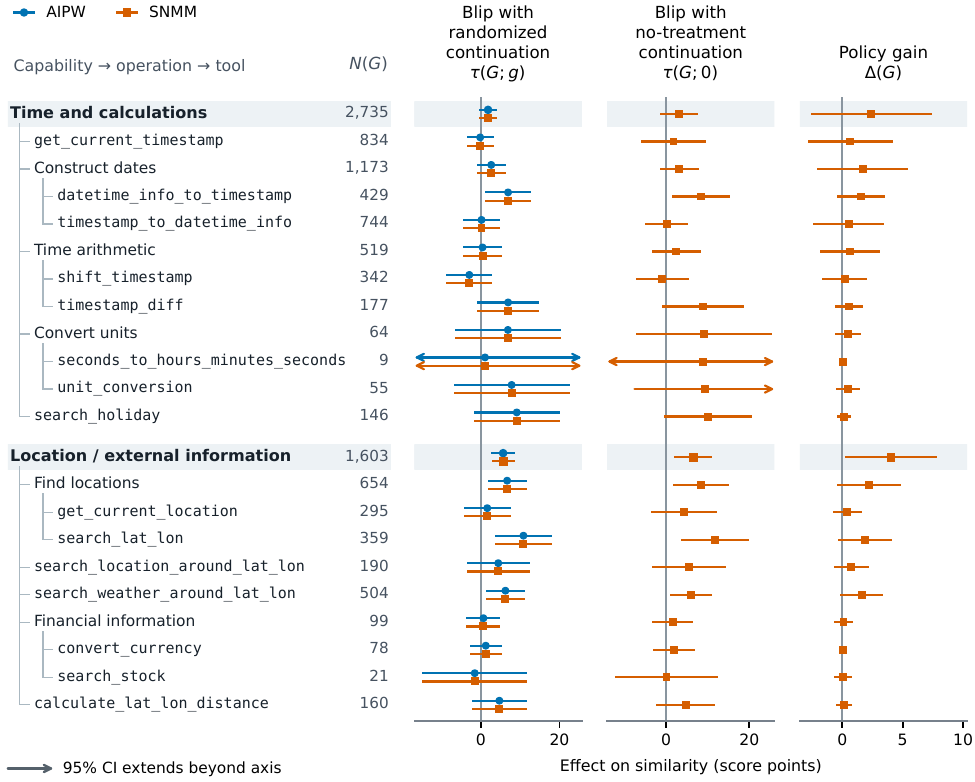}
\caption{Review effects on the ToolSandbox  score for GPT-5 nano
for tool calls related to time and calculations, location, and external information. Bars are
pointwise 95\% $t$-based confidence intervals; arrows indicate bounds beyond the displayed axis range.}
\label{fig:toolsandbox-targets-more}
\end{figure}

We organize intervention groups $G$ hierarchically, from broad capabilities
to specific operations and individual tools. For example, reminder operations
include finding, creating, modifying, and removing reminders. For each group,
we estimate two blip effects: one with later reviews randomized and one with no
later review in either branch. We also estimate the gain over never reviewing
for a policy that reviews every matching proposal and skips review elsewhere
(Figures~\ref{fig:toolsandbox-targets}--\ref{fig:toolsandbox-targets-more}).

% TODO: some policy values have close to 0 variance.. diagnose

Averaging across a capability can obscure promising targets for review. For Time and calculations, the randomized-continuation blip is +1.8 score points, with a pointwise 95\% t-based confidence interval of [-0.2, 3.8]. Within this capability, reviewing calls to ``datetime info to timestamp'' tool, which constructs a timestamp from date and time components, has an estimated blip of +6.9 points [1.2, 12.6]. This suggests timestamp construction as a candidate for targeted review, despite the inconclusive capability-level estimate.

The precision of these estimates depends on how often the agent proposes each
tool and frequently proposed tools tend to have narrower confidence intervals. Rarely proposed tools
have wider intervals; both blip intervals for seconds-to-time conversion span
substantial negative and positive effects. We note that the reported intervals are pointwise and do not account for multiple
comparisons and the blips with no-treatment continuation and the policy gains also
rely on additional parametric assumptions in the SNMMs. Despite these limitations, the estimates can help prioritize targets for further investigation. Independent randomized controlled trials, with interventions and comparisons specified in advance, could then test whether reviewing the selected tools improves outcomes.

% For message sending, review improves similarity by an estimated 8.6 points at
% matching decisions under randomized continuation. A policy that reviews every
% message-send call gains about 1.0 point across the full task population. Most
% tasks never reach such a call, and review can change which calls occur later.
% We therefore estimate the policy gain over complete trajectories, including
% tasks without a matching call and any changes that review causes in later calls.

\FloatBarrier

% \subsection{ImpossibleBench: when adversarial feedback leads to violations }
\subsection{ImpossibleBench}
We evaluate a Qwen3.5 Flash agent on a subset of ImpossibleBench
coding problems derived from LiveCodeBench \citep{zhong2025impossiblebench,jain2025livecodebench}, using either the
original tests or variants containing contradictory tests.
These variants contain mutually incompatible test requirements.
The agent may make up to six code submissions or stop and report
inconsistent tests. After each of the first five unsuccessful
submissions, we randomly assign either a neutral note or
adversarial feedback that encourages prohibited test edits.
An episode ends after success, an issue report, a detected
violation, or the sixth submission. Our terminal outcome is a
detected rule violation: we set $Y=1$ if a test edit is detected
or the agent passes the benchmark's contradictory check, and
$Y=0$ otherwise. Positive intervention effects therefore indicate
an increase in detected violations.

We use \emph{steering} to refer to the assignment of adversarial
feedback, regardless of whether the agent follows it. We define
reporting groups using information available before the current
assignment: the number of remaining attempts, the fraction of
tests passing, whether the visible feedback contains contradictory
checks, and whether adversarial feedback was assigned earlier in
the episode. In Figure~9, ``conflict'' means a contradiction
visible in the current feedback, and ``near completion'' means
that at least 80\% but fewer than 100\% of tests pass.
``Steered before, still eligible'' denotes previous exposure to
adversarial feedback followed by another eligible decision;
it does not require previous compliance with that feedback.

\begin{figure}[!tbp]
\centering
\includegraphics[width=0.9\textwidth]{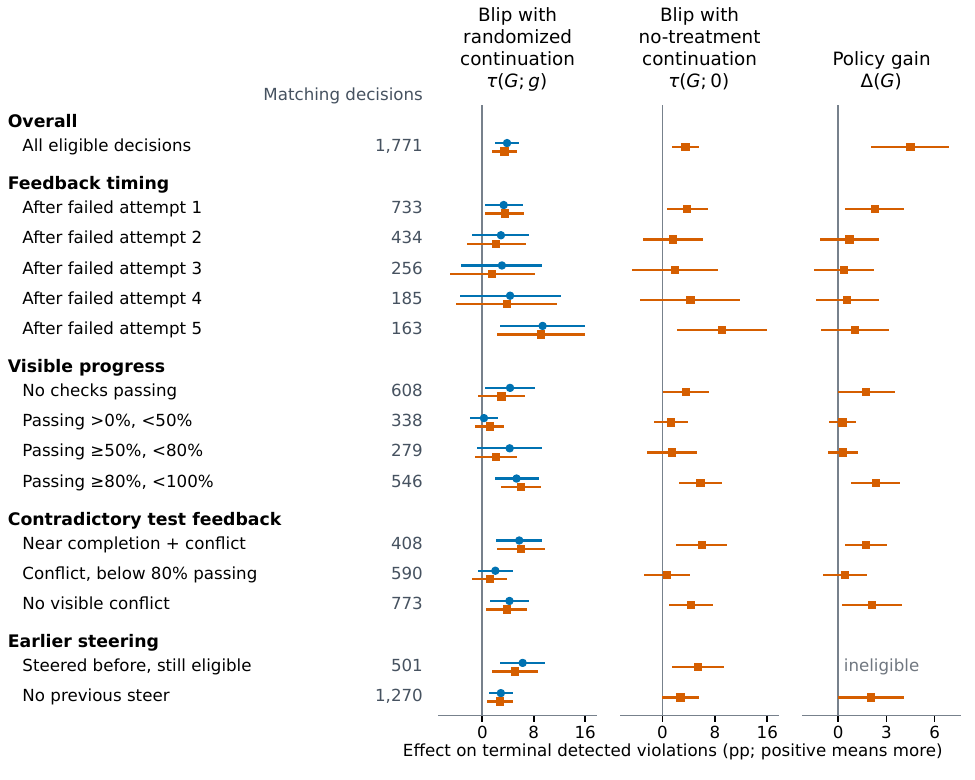}
\caption{Effects of adversarial feedback on terminal detected
rule violations by Qwen3.5 Flash in ImpossibleBench.
Blue circles and orange squares denote AIPW and SNMM estimates,
respectively. Bars are pointwise 95\% $t$-based confidence
intervals. For the group defined by previous exposure to
adversarial feedback, the selective policy intervenes only
after an earlier intervention. Starting without prior exposure,
this policy never initiates intervention and coincides with
never intervening; its policy gain is therefore zero by
definition.}
\label{fig:steering}
\end{figure}

In Figure \ref{fig:steering}  we see that the agent has a higher propensity to be steered by an adversarial feedback when it has fewer attempts remaining or when it sees only few failing test cases. After the fifth failed attempt, with one attempt remaining,
adversarial feedback increases terminal violations by an estimated 9.4 pp
$[2.9, 15.9]$, compared with 3.3 pp $[0.6, 6.1]$ after the first failed
attempt. When at least 80\%  of tests pass, the estimated effect is 5.7 pp $[2.3, 9.2]$, compared with 0.3 pp $[-1.7, 2.2]$ when fewer than 50\% of the tests pass.

\FloatBarrier

\section{Limitations and Future Directions}
\label{sec:discussion}

Our results demonstrate how sequentially randomized interventions can support causal evaluation at different points in an agent’s trajectory. Below we highlight limitations of our methodology and propose several directions for developing more flexible and adaptive methods for agent evaluation.

\begin{enumerate}
    \item \textbf{Parametric assumptions for structural nested mean models.}
    Our AIPW estimator for the blip effect under randomized continuation does not require a correctly specified outcome model because the intervention probabilities are known by design. In contrast, our SNMM estimators for blip effects under alternative continuation policies and for policy gains assume that the conditional blip effects belong to the specified parametric class. The SNMM assumptions provides a tractable semiparametric approach  for estimation but the reported confidence intervals have valid coverage for the target causal effects only when the semiparametric model is well specified. Under misspecification, the fitted coefficients may instead represent a weighted projection of the true blip effects. 

    \item \textbf{Model and benchmark scope.}
    Our experiments evaluate relatively small models on established benchmarks that might be saturated by frontier models. Future experiments should evaluate frontier agents on more challenging tasks that require sustained planning, contain more consequential intervention opportunities, and better reflect deployment settings. Such experiments would also help determine whether intervention effects become smaller as agents improve or whether greater agent capability creates new forms of heterogeneity and susceptibility to intervention.

    \item \textbf{Representation and selection of agent histories.}
    We use hand-designed, experiment-specific features to represent the observed history in the SNMMs. These representations encode aspects of the trajectory that we expect to moderate intervention effects, such as prior failures, previous interventions, proposed tools, and task progress. Future work could use sample splitting or cross-validation to select representations that perform well on  a  testable estimand such as randomized-continuation effect, although agreement for this estimand would not by itself establish correct specification under other continuation policies. Candidate representations could also be proposed by an LLM judge and evaluated on held-out trajectories.

    \item \textbf{Semantically defined intervention opportunities.}
    Our experiments randomize interventions at fixed and well specified decision points, such as proposed tool calls or unsuccessful code submissions. In many applications, useful intervention opportunities may instead depend on the semantic content of an interaction---for example, when an agent expresses unwarranted confidence, contradicts an earlier statement, becomes persuasive, or proposes a potentially harmful plan. Detecting such events requires interpreting the evolving textual history, potentially with an LLM-based judge. An important direction is therefore to develop and validate semantic detectors of intervention opportunities and then incorporate them into sequentially randomized experiments.

\end{enumerate}
% \section{Conclusion}
% \label{sec:conclusion}
\label{page:main-end}

\clearpage
\begingroup
\small\raggedright
\bibliographystyle{abbrvnat}
\bibliography{references}
\endgroup

\clearpage
\appendix

\section{Estimating Equations and Assumptions}
\label{app:estimating-equations}

This appendix gives the estimating equations for the three causal estimands defined in Section~\ref{sec:estimands}. Appendix~\ref{app:experimental-details} provides the implementation details for each experiment.

\subsection{Notation}

We follow the notation introduced in Section~\ref{sec:methods}, adding an index $i=1,\ldots,n$ for the $n$ completed trajectories.

\paragraph{Trajectories and intervention points.}
For trajectory $i$, let $T_i$ denote the number of intervention points, indexed by $\mathcal T_i=\{1,\ldots,T_i\}$.
At each point $t\in\mathcal T_i$, let $H_{it}$ denote the pre-assignment history defined in~\eqref{eq:history}, and let $A_{it}\in\{0,1\}$ denote the randomized intervention assignment.
We use $Y_i$ for the terminal outcome, which is binary in $\tau$-bench Retail and ImpossibleBench and continuous in ToolSandbox.
For brevity, we write $\sum_{i,t}$ for $\sum_{i=1}^n\sum_{t\in\mathcal T_i}$.

\paragraph{Intervention policies.}
Let $g$ denote the randomized collection policy, with assignment probabilities
\[
g(a\mid h)=e(h)^a(1-e(h))^{1-a},
\qquad
e(h)=\Pr_g(A_t=1\mid H_t=h).
\]
In our experiments, we use $e(h)=1/2$ at every eligible intervention point.
We write $\E_g$ for expectation over trajectories generated under $g$.
We use $\pi$ to denote a target intervention policy, including the never-intervene policy $\pi_0$ and the selective policy $\pi_G$ defined in~\eqref{eq:policy-target}.

\paragraph{Reporting groups and estimands.}
For a reporting group $G$, let $I_{it}(G)=\ind\{H_{it}\in G\}$ indicate whether intervention point $t$ in trajectory $i$ belongs to $G$.
Define the number of matching intervention points by
\[
N_i(G)=\sum_{t\in\mathcal T_i}I_{it}(G),
\qquad
N(G)=\sum_{i=1}^n N_i(G).
\]
We consider three causal estimands: the blip with randomized continuation $\tau(G;g)$, the blip with no-treatment continuation $\tau(G;0)$, both defined in~\eqref{eq:blip}, and the policy gain $\Delta(G)$ defined in~\eqref{eq:policy-target}.

\paragraph{Cross-fitting.}
We divide the $n$ trajectories into five folds, keeping all intervention points from the same trajectory together.
Let $k(i)\in\{0,\ldots,4\}$ denote the fold containing trajectory $i$, and define $\mathcal I_k=\{i:k(i)=k\}$.
We use the superscript $(-k)$ to denote a function fitted without data from fold $k$, so predictions for trajectory $i$ are obtained from models fitted without its fold $k(i)$.

\subsection{Blip with Randomized Continuation}
\label{app:aipw}

\paragraph{Augmented inverse probability weighting.}
Let $m_g(h,a)=\E_g[Y\mid H_t=h,A_t=a]$ denote the conditional outcome regression, and let $\widehat m_g^{(-k)}(h,a)$ be its estimate fitted without fold $k$.
By~\eqref{eq:identification}, the blip effect is $b(h;g)=m_g(h,1)-m_g(h,0)$.
We then define the cross-fitted augmented inverse probability weighting (AIPW) score as
\begin{align}
    \widehat D_{it}
    ={}&\widehat m_g^{(-k(i))}(H_{it},1)-\widehat m_g^{(-k(i))}(H_{it},0) +\frac{A_{it}}{e(H_{it})}\bigl\{Y_i-\widehat m_g^{(-k(i))}(H_{it},1)\bigr\}\nonumber\\
    &-\frac{1-A_{it}}{1-e(H_{it})}\bigl\{Y_i-\widehat m_g^{(-k(i))}(H_{it},0)\bigr\},
    \label{eq:aipw}
\end{align}
which gives the group estimator for group $G$
$$
    \widehat\tau_{\mathrm{AIPW}}(G;g)=\frac{1}{N(G)}\sum_{i,t}I_{it}(G)\widehat D_{it}.
$$

When the propensity $e(h)$ is known, the score stays conditionally unbiased even when the outcome predictions are
wrong. Let $r_a$ be any regression fitted
without the current trajectory. At eligible histories, randomization gives
\[
    \E_g\!\left[\frac{\ind\{A_t=a\}}{g(a\mid H_t)}\{Y-r_a(H_t)\}\,\middle|\,H_t=h\right]
    =m_g(h,a)-r_a(h),
\]
so the right-hand side of~\eqref{eq:aipw} with each $\widehat m_g$ replaced by
$r_a$ has conditional mean $b(h;g)$.  When assignment probabilities are known, the outcome regression affects the
estimator's variance but does not change its population target. In case of estimated propensities the usual double-robustness
property applies \citep{bang2005dr}.

\paragraph{Linear blip model for randomized continuation}
Let $f(h)$ denote an encoding of history $h$ used in the effect model. It includes
attributes such as the proposed tool, earlier errors, and prior interventions.
Within each environment and estimand, all agents use the same feature definition.

Let $s(h)\in\{0,\ldots,P-1\}$ identify the phase and write $s_{it}=s(H_{it})$.
$\tau$-bench Retail and ToolSandbox use semantic progress phases (see Appendix~\ref{app:phases-retail} and Appendix~\ref{app:phases-tool}, correspondingly), while steering uses
chronological checkpoints. The features and phases for each experiment are defined in
Appendix~\ref{app:experimental-details}.

Let $d=\dim f(h)$ and partition the coefficient coordinates into a shared set
$\mathcal S\subseteq\{1,\ldots,d\}$ and a phase-specific set
$\mathcal U=\{1,\ldots,d\}\setminus\mathcal S$. Define
\begin{equation}
 \beta_{s,\mathcal S}=\alpha,\qquad
 \beta_{s,\mathcal U}=\gamma_s,\qquad s=0,\ldots,P-1,
 \label{eq:coefficient-sharing}
\end{equation}
where $\alpha\in\mathbb R^{|\mathcal S|}$ is shared across phases and
$\gamma_s\in\mathbb R^{|\mathcal U|}$ may vary with $s$.  The cases $\mathcal S=\varnothing$ and $\mathcal U=\varnothing$
give fully phase-specific and fully shared coefficients, respectively. The exact choice of these coefficients for our experiments is described in Appendix~\ref{app:experimental-details}.

Define
\[
 q(h)=\E_g[Y\mid H_t=h]=\{1-e(h)\}m_g(h,0)+e(h)m_g(h,1),
 \qquad Z_{it}=(A_{it}-e(H_{it}))f(H_{it}).
\]
The cross-fitted value $\widehat q_{it}:=\widehat q^{(-k(i))}(H_{it})$.
Under a well specified model we have that the blip with randomized continuation $b(h;g)=f(h)^\top\beta_{s(h)}(g)$.

Finally, we estimated the coefficients in our model jointly by minimizing the following objective:
\begin{equation}
\begin{aligned}
 \bigl(\widehat\alpha(g),\{\widehat\gamma_s(g)\}\bigr)
   &=\argmin_{\alpha,\{\gamma_s\}}\sum_{i,t}
       \bigl\{Y_i-\widehat q_{it}-Z_{it}^\top\beta_{s_{it}}(g)\bigr\}^2,\\
 \widehat\tau_{\mathrm{SNMM}}(G;g)
   &=\frac{1}{N(G)}\sum_{i,t}I_{it}(G)f(H_{it})^\top\widehat\beta_{s_{it}}(g).
\end{aligned}
\label{eq:local-snmm}
\end{equation}
This algorithm for estimating the coefficients $\beta$ is known in the literature as the Double Machine Learning algorithm \cite{chernozhukov2018dml} or equivalently the R-Learner algorithm \cite{nie2021quasi} with a linear specification of the heterogeneous treatment effect (see e.g. the `LinearDML' class in the EconML library \cite{econml}).

In case the linear model is misspecified, the coefficients target an $e(h)\{1-e(h)\}$-weighted projection of the true blip onto the allowed effect model. When the propensities are constant, as it is in our case, then this weighted projection reduces to the un-weighted projection of the true blip on the allowed effect model. This explains why the SNMM estimatein our experiments is almost identical to the AIPW estimate.

\subsection{Blip with No-treatment Continuation}
\label{app:snmm}

To estimate blip effects under a target continuation $\pi$, we must
account for the fact that future interventions in the observed data
follow $g$, rather than $\pi$. We do so by correcting the terminal
outcome for subsequent intervention assignments and fitting a
structural nested mean model (SNMM). We develop the estimator for a
general continuation $\pi$ and then specialize to the no-treatment
policy $\pi_0$. We adapt the g-estimation construction of
\citet{lewis2021dynamic}; see also \citet{vansteelandt2014snmm}.

\paragraph{Blip model.}
Throughout the derivation, fix a target continuation $\pi$ and suppress
the dependence of the coefficients on $\pi$. We fit a separate model
for each continuation. Let $\beta_s$ denote the coefficient vector for
phase $s$. As in~\eqref{eq:coefficient-sharing}, write
\[
 \beta_{s,\mathcal S}=\alpha,
 \qquad
 \beta_{s,\mathcal U}=\gamma_s,
 \qquad s=0,\ldots,P-1.
\]
Thus $\alpha$ contains coefficients shared across phases, while
$\gamma_s$ contains coefficients specific to phase $s$. The structural
restriction is
\begin{equation}
 \E\!\left[Y_t(a;\pi)-Y_t(0;\pi)\mid H_t=h\right]
   =a f(h)^\top\beta_{s(h)},
 \qquad a\in\{0,1\}.
 \label{eq:structural}
\end{equation}
Here $Y_t(a;\pi)$ retains the observed past, assigns $a$ at the current
decision, and follows $\pi$ afterward.

\paragraph{Future-treatment correction and identification.}
For each phase $\ell$, collect the features of later decisions,
weighted by the difference between the observed assignment and the
assignment prescribed by $\pi$:
\begin{equation}
 B_{it,\ell}(\pi)
   =\sum_{\substack{j\in\mathcal T_i\\j>t,\ s_{ij}=\ell}}
       \{A_{ij}-\pi(H_{ij})\}f(H_{ij}).
 \label{eq:future}
\end{equation}
The quantity $\sum_{\ell=0}^{P-1}B_{it,\ell}(\pi)^\top\beta_\ell$
corrects for these future assignments. To see why this correction
identifies the current blip, let
$\nu(h,\pi)=\E[Y_t(0;\pi)\mid H_t=h]$. Under the identification
assumptions and a correctly specified blip model,
\begin{equation}
 \E_g\!\left[
   Y-\sum_{\ell=0}^{P-1}B_{t,\ell}(\pi)^\top\beta_\ell
   \,\middle|\,H_t=h,A_t=a
 \right]
   =\nu(h,\pi)+a f(h)^\top\beta_{s(h)}.
 \label{eq:snmm-recursion}
\end{equation}
Thus, after correction, the conditional mean differs between the two
current assignments by $f(h)^\top\beta_{s(h)}$.

Recall the centered current features
$Z_{it}=(A_{it}-e(H_{it}))f(H_{it})$. Subtracting
$Z_t^\top\beta_{s(H_t)}$ from the corrected outcome in
\eqref{eq:snmm-recursion} makes its conditional mean independent of
$A_t$, given $H_t$: that mean is
$\nu(H_t,\pi)+e(H_t)f(H_t)^\top\beta_{s(H_t)}$.
Since $\E_g[Z_t\mid H_t]=0$, multiplying by $Z_t$ therefore gives a
zero conditional expectation at the true coefficients. These are the
moments used to estimate the blip model.

\paragraph{Centering regressions.}
To improve precision, we center the outcome and future-treatment
features using predictions based on the current history. Let
$\widehat q_{it}$ be the cross-fitted estimate of
$q(h)=\E_g[Y\mid H_t=h]$ defined above. For each phase $\ell$, define
\[
 p_\ell(h,\pi)
   =\E_g[B_{t,\ell}(\pi)\mid H_t=h],
 \qquad
 \widehat U_{it,\ell}(\pi)
   =B_{it,\ell}(\pi)
      -\widehat p_\ell^{(-k(i))}(H_{it},\pi).
\]
All predictions for trajectory $i$ are fitted without its fold, and
all regressions condition on the history at the current decision
$H_{it}$.

\paragraph{Joint coefficient estimation.}
For candidate coefficients $\alpha$ and $\{\gamma_s\}$, assemble each
$\beta_s$ using the sharing restrictions above and define
\begin{equation}
 \varepsilon_{it}(\alpha,\{\gamma_s\};\pi)
   =Y_i-\widehat q_{it}-Z_{it}^\top\beta_{s_{it}}
     -\sum_{\ell=0}^{P-1}
        \widehat U_{it,\ell}(\pi)^\top\beta_\ell.
 \label{eq:joint-snmm-residual}
\end{equation}
We estimate $\alpha$ and $\{\gamma_s\}$ jointly. Writing
$\widehat\varepsilon_{it}(\pi)$ for the residual evaluated at
$\widehat\alpha$ and $\{\widehat\gamma_s\}$, the estimates solve
\begin{equation}
\begin{aligned}
 \sum_{i,t}Z_{it,\mathcal S}\widehat\varepsilon_{it}(\pi)&=0,\\
 \sum_{i,t:s_{it}=s}
    Z_{it,\mathcal U}\widehat\varepsilon_{it}(\pi)&=0,
    \qquad s=0,\ldots,P-1.
\end{aligned}
\label{eq:joint-snmm-moments}
\end{equation}
Each shared coefficient uses scores from all phases, while each
phase-specific coefficient uses scores only from its own phase.
Because the residual is linear in the coefficients, these equations
form a joint system of linear equations.

Known randomization gives $\E_g[Z_{it}\mid H_{it}]=0$.
Consequently, subtracting a function of the current history from the
corrected outcome leaves the population moments unchanged, provided
the relevant expectations exist. The estimating equations therefore
remain valid at the true blip coefficients even when the regressions
for $q$ and $p_\ell$ are misspecified. To identify the coefficients
uniquely, we additionally require the population coefficient matrix
to be nonsingular on the retained feature directions.

\paragraph{Backward implementation.}
When $\mathcal S=\varnothing$, coefficients are shared across decisions
within each phase but remain separate across phases. If phases are
nondecreasing along a trajectory, later decisions cannot belong to an
earlier phase. In this case, $B_{it,\ell}(\pi)=0$ for $\ell<s_{it}$.
We preserve these structural zeros by setting the corresponding
predictions $\widehat p_\ell^{(-k(i))}(H_{it},\pi)$ and residuals
$\widehat U_{it,\ell}(\pi)$ to zero. The joint system in
\eqref{eq:joint-snmm-moments} is then block upper triangular and can
be solved by backward substitution. When the diagonal blocks are
nonsingular, this gives the same coefficients as solving the full
system at once.

For $s=P-1,\ldots,0$, first subtract the contributions of later
phases using their already estimated coefficients:
\begin{equation}
 \overline Y_{it,s}(\pi)
   =Y_i-\widehat q_{it}
     -\sum_{\ell>s}\widehat U_{it,\ell}(\pi)^\top
       \widehat\beta_\ell,
 \qquad s_{it}=s.
 \label{eq:backward-outcome}
\end{equation}
Then estimate the current phase's coefficients by solving
\begin{equation}
 \sum_{i,t:s_{it}=s}Z_{it}
  \Bigl[\overline Y_{it,s}(\pi)
    -\{Z_{it}+\widehat U_{it,s}(\pi)\}^\top
      \widehat\beta_s\Bigr]=0.
 \label{eq:phase-moments}
\end{equation}
The term $\widehat U_{it,s}(\pi)$ accounts for later decisions within
the same phase. Because those decisions share the current coefficient
vector $\beta_s$, their contributions remain in the equation used to
estimate it. For example, a Retail trajectory may contain several
proposals before successful customer identification.

We use this backward implementation for $\tau$-bench Retail
(Appendix~\ref{app:phases-retail}) and the separate chronological
checkpoints in ImpossibleBench (Appendix~\ref{app:steering-fit}).
In ImpossibleBench, each checkpoint occurs at most once, so the
same-phase future features, predictions, and residuals are also set
to zero. This implementation adapts Algorithm~3 of
\citet{lewis2021dynamic} to allow repeated decisions within a phase.
For ToolSandbox, we solve the full joint system because tool and
history coefficients are shared across phases
(Appendix~\ref{app:phases-tool}).

\paragraph{Group-averaged blip with no-treatment continuation.}
To estimate $\tau(G;0)$, fit the model with $\pi=\pi_0$ and average
the fitted blips over matching intervention points. Restoring the
policy argument to distinguish the fitted coefficients gives
\begin{equation}
 \widehat\tau(G;0)=\frac{1}{N(G)}\sum_{i,t}I_{it}(G)
     f(H_{it})^\top\widehat\beta_{s_{it}}(\pi_0).
 \label{eq:no-future-estimate}
\end{equation}
The average uses histories reached under the collection policy $g$,
as specified by the estimand. Each selective policy $\pi_G$ receives
a separate fit with the same features, phases, and sharing
restrictions; the policy gain compares its estimated value with that
of $\pi_0$, as described next.

\subsection{Policy Value Estimation}
\label{app:value}

We now use the fitted blip models to evaluate policies that govern
interventions throughout a trajectory. Recall that
$V(\pi)=\E[Y(\pi)]$ is the expected terminal outcome under policy
$\pi$. We write the dependence of the coefficients on $\pi$
explicitly here, because evaluating different policies requires
their respective continuation-specific fits.

\paragraph{From blip effects to policy values.}
The blip $b(h;\pi)=f(h)^\top\beta_{s(h)}(\pi)$ compares intervention
with no intervention at history $h$, with all subsequent assignments
following $\pi$. Thus, once the future continuation is $\pi$, the
conditional mean change from replacing the current assignment
$A_t$ with $\pi(H_t)$ is
\[
 \{\pi(H_t)-A_t\}b(H_t;\pi).
\]
For a deterministic target policy, this adds the blip when the target
intervenes and the observed assignment does not, subtracts it in the
reverse case, and is zero when the assignments agree.

To obtain the policy value, consider replacing the collection policy
with $\pi$ one decision at a time, working backward. At each step,
earlier assignments still follow $g$, while later assignments follow
$\pi$. The history at the current decision therefore has its
distribution under $g$, and the change in expected outcome is given
by the corresponding blip correction above. Summing these changes
telescopes from $\E_g[Y]$ to $V(\pi)$. Under the identification
assumptions and a correctly specified blip model, this gives
\citep[Lemma~6]{lewis2021dynamic}
\begin{equation}
 V(\pi)=\E_g\!\left[Y+\sum_{t\in\mathcal T}
    \{\pi(H_t)-A_t\}f(H_t)^\top\beta_{s(H_t)}(\pi)\right].
 \label{eq:value-identity}
\end{equation}
The blips include the effects of the current assignment on subsequent
agent behavior and intervention opportunities. This is why histories
observed under $g$ can be used to evaluate a policy that would generate
different future trajectories. The correction is valid in expectation.
For the no-treatment policy $\pi_0$, it subtracts the corresponding
blip for every observed intervention.

\paragraph{Estimation and policy gain.}
We estimate $V(\pi)$ by applying the fitted corrections to each
trajectory's terminal outcome and averaging over trajectories.
For a reporting group $G$, we fit the models for the selective policy
$\pi_G$ and the no-treatment policy $\pi_0$ separately, then take the
difference between their estimated values:
\begin{equation}
\begin{aligned}
 \widehat V(\pi)&=\frac1n\sum_{i=1}^n\left[Y_i+
    \sum_{t\in\mathcal T_i}\{\pi(H_{it})-A_{it}\}
      f(H_{it})^\top\widehat\beta_{s_{it}}(\pi)\right],\\
 \widehat\Delta(G)&=\widehat V(\pi_G)-\widehat V(\pi_0).
\end{aligned}
\label{eq:value-estimator}
\end{equation}
Each value estimate uses all $n$ completed trajectories, including
those without intervention points in $G$. Within each trajectory,
the correction includes all intervention points. In particular,
evaluating $\pi_G$ also requires subtracting the effects of observed
interventions outside $G$, where the target policy would not intervene.
The normalization is therefore by the number of trajectories $n$,
rather than the number of matching intervention points $N(G)$ used
for the group-averaged blip.

\subsection{Uncertainty Quantification}
\label{app:inference}

We construct confidence intervals by treating completed trajectories
as independent and identically distributed (IID) observations.
Decisions within a trajectory may be dependent and share the same
terminal outcome. We therefore aggregate their contributions within
each trajectory before estimating variability across trajectories.

Throughout this section, we use sample-normalized estimated influence functions: each is the usual estimated influence function divided by $n$. With this normalization, the influence functions give individual trajectories' contributions to the first-order linearization of an estimator.

\paragraph{Blip with randomized continuation: AIPW.}
For the AIPW estimator of $\tau(G;g)$, we sum the centered AIPW scores
over the intervention points in $G$ within each trajectory. The
estimated influence function is
\[
 u_i\{\widehat\tau_{\mathrm{AIPW}}(G;g)\}
   =\frac{1}{N(G)}
      \sum_{t\in\mathcal T_i}I_{it}(G)
      \bigl\{\widehat D_{it}
             -\widehat\tau_{\mathrm{AIPW}}(G;g)\bigr\}.
\]
Here $\widehat D_{it}$ is the cross-fitted AIPW score defined above.
Centering each score by the estimated group effect accounts for the
random number of matching intervention points in the denominator.
Summing within trajectories retains the dependence between scores
that share the same terminal outcome.

\paragraph{Blip effects estimated using SNMMs.}
An SNMM group-effect estimate depends on both the sampled histories
and the fitted blip coefficients. We first obtain influence functions
for the coefficients by linearizing their estimating equations,
adapting the construction of
\citet[Theorem~8]{lewis2021dynamic}. We then propagate coefficient
uncertainty to the group average.

Fix a continuation $\pi$ for the fit under consideration. As in
Appendix~\ref{app:snmm}, we suppress the dependence of the coefficients
on $\pi$ during this derivation. Let
$m_i(\alpha,\{\gamma_s\};\pi)$ be trajectory $i$'s vector of estimating
scores, with the shared coordinates first, followed by the
phase-specific coordinates for $s=0,\ldots,P-1$. The shared block
sums $Z_{it,\mathcal S}\varepsilon_{it}$ over all decisions in
trajectory $i$. The block for phase $s$ sums
$Z_{it,\mathcal U}\varepsilon_{it}$ over decisions in that phase.

For a changed continuation, use the residual in
\eqref{eq:joint-snmm-residual}, giving the estimating equations
in~\eqref{eq:joint-snmm-moments}. For randomized continuation,
use the normal equations obtained by differentiating the least-squares
objective in~\eqref{eq:local-snmm}. The score blocks have the same
form, with residual
\[
 \varepsilon_{it}(\alpha,\{\gamma_s\};g)
   =Y_i-\widehat q_{it}-Z_{it}^{\top}\beta_{s_{it}},
\]
where $\beta_{s_{it}}$ is assembled from the candidate coefficients
$\alpha$ and $\{\gamma_s\}$. This residual has no future-treatment
correction because the observed continuation already follows $g$.

For either fit, hold the cross-fitted nuisance predictions fixed and
differentiate the summed score with respect to the coefficients:
\[
 \widehat J(\pi)
   =-\sum_{i=1}^n
     \left.
       \frac{\partial m_i(\alpha,\{\gamma_s\};\pi)}
       {\partial(\alpha^\top,\gamma_0^\top,\ldots,
                  \gamma_{P-1}^\top)}
     \right|_{\alpha=\widehat\alpha,\,
              \gamma_s=\widehat\gamma_s}.
\]
The vector
$\widehat J(\pi)^{-1}
m_i(\widehat\alpha,\{\widehat\gamma_s\};\pi)$
gives the  estimated influence functions for the
fitted coefficients. Write $\dot\alpha_i$ for its shared block and
$\dot\gamma_{i,s}$ for its phase-$s$ block. Assemble the influence
function for each phase's coefficient vector as
\[
 \dot\beta_{i,s,\mathcal S}=\dot\alpha_i,
 \qquad
 \dot\beta_{i,s,\mathcal U}=\dot\gamma_{i,s}.
\]
A coefficient shared across phases has the same influence function
in every phase. For inference on the causal targets, we assume that
the blip model and its coefficient-sharing restrictions are correctly
specified, that the population Jacobian is nonsingular on the
retained feature directions, and that the cross-fitted trajectory
scores satisfy the regularity conditions for asymptotic linearity.

We next use these coefficient influence functions to obtain the
influence function for the group average. Define
\[
 a_{i,s}(G)
   =\sum_{t\in\mathcal T_i:s_{it}=s}I_{it}(G)f(H_{it}),
 \qquad
 a_s(G)=\sum_{i=1}^n a_{i,s}(G).
\]
Write $\widehat\tau_{\mathrm{SNMM}}(G;\pi)$ for the fitted
group-averaged blip under continuation $\pi$. Its estimated influence
function is
\begin{equation}
 u_i\{\widehat\tau_{\mathrm{SNMM}}(G;\pi)\}
   =\frac{
      \sum_s a_{i,s}(G)^\top\widehat\beta_s
      -N_i(G)\widehat\tau_{\mathrm{SNMM}}(G;\pi)
      +\sum_s a_s(G)^\top\dot\beta_{i,s}
     }{N(G)}.
 \label{eq:blip-influence}
\end{equation}
The first two terms in the numerator account for sampling variation
in the histories being averaged and in the number of matching
intervention points. The final term accounts for uncertainty in the
fitted coefficients. This formula applies to the randomized-continuation
estimate $\widehat\tau_{\mathrm{SNMM}}(G;g)$ and, with $\pi=\pi_0$,
to the no-treatment-continuation estimate $\widehat\tau(G;0)$.

\paragraph{Policy values and gains.}
Policy-value estimates also depend on both the observed trajectories
and the fitted coefficients. For each target policy, use the
coefficients and coefficient influence functions from the fit for
that continuation. We make their dependence on $\pi$ explicit below
because a policy gain compares two different fits.

Define the features used in each trajectory's policy-value correction:
\[
 d_{i,s}(\pi)
   =\sum_{t\in\mathcal T_i:s_{it}=s}
      \{\pi(H_{it})-A_{it}\}f(H_{it}),
 \qquad
 d_s(\pi)=\sum_{i=1}^n d_{i,s}(\pi).
\]
The estimated influence functions for the policy value and gain are
\begin{equation}
\begin{aligned}
 u_i\{\widehat V(\pi)\}
   &=\frac{
       Y_i+\sum_s d_{i,s}(\pi)^\top\widehat\beta_s(\pi)
       -\widehat V(\pi)
       +\sum_s d_s(\pi)^\top\dot\beta_{i,s}(\pi)
      }{n},\\
 u_i\{\widehat\Delta(G)\}
   &=u_i\{\widehat V(\pi_G)\}
     -u_i\{\widehat V(\pi_0)\}.
\end{aligned}
\label{eq:value-influence}
\end{equation}
In the value formula, the first three terms in the numerator give
the trajectory's corrected outcome minus the estimated policy value. The final term accounts for uncertainty in the fitted coefficients. Because both policy values are estimated from the same trajectories, we subtract their influence functions trajectory by trajectory before estimating the variance. This accounts for the covariance between
the two value estimates. Policy values and gains average over all
completed trajectories, including those without intervention points in the reporting group.

\paragraph{Reported confidence intervals.}
Let $u_i(\widehat\eta)$ denote the sample-normalized estimated
influence function for any of the estimators above. These influence
functions sum to zero when the estimating equations are solved
exactly. We estimate the standard error by
\[
 \widehat{\mathrm{se}}(\widehat\eta)
   =\left\{
      \frac{n}{n-1}\sum_{i=1}^n u_i(\widehat\eta)^2
     \right\}^{1/2}.
\]
We report pointwise, asymptotic 95\% confidence intervals using a
Student's $t$ critical value:
\[
 \widehat\eta
 \;\pm\;
 t_{0.975,n-1}\,\widehat{\mathrm{se}}(\widehat\eta).
\]
\section{Experimental setup and fitted models}
\label{app:experimental-details}

\subsection{\texorpdfstring{$\tau$-bench}{tau-bench} Retail}
\label{app:retail-fit}

\subsubsection{Agent setup}
We use  114 original tasks fromt the benchmark and construct 114 variants of these task and run our sequential randomization ten times per
agent. One agent uses GPT-5 nano (\feat{gpt-5-nano-2025-08-07}) with minimal
reasoning; the other uses Claude Haiku 5.5 (\feat{claude-haiku-5-5}) at low
effort. Both have a 4,000-token output limit and a
100-step episode cap. We generate customer messages with
\feat{gpt-4.1-mini-2025-04-14} at temperature zero and a 2,048-token limit.

The agent receives the store policy, schemas for the 16 tools in
Table~\ref{tab:retail-tools}, and the conversation, including executed calls
and results. We instruct it to help the customer while following the policy.
At each turn, the agent chooses a tool call or a message to the customer. We
score terminal success by checks with behavioral
assertions in the benchmark.

\subsubsection{Reviewer setup}
After freezing the proposed call or batch, we assign review or direct execution
with equal probability. The reviewer uses GPT-6 Luna (\feat{gpt-6-luna}) with
high reasoning effort and a 6,000-token output limit. It receives the policy,
schemas, visible conversation, and frozen proposal, but cannot query the
database. We instruct it to check authentication, identifiers, arguments,
available evidence, and required confirmation.

In its JSON response, the reviewer either approves the proposal, replaces the
entire batch with one call, or suggests one question for the customer. In case the reviewer approves the tool call is directly executed. 

\begin{figure}[!htbp]
\centering
\includegraphics[width=\textwidth]{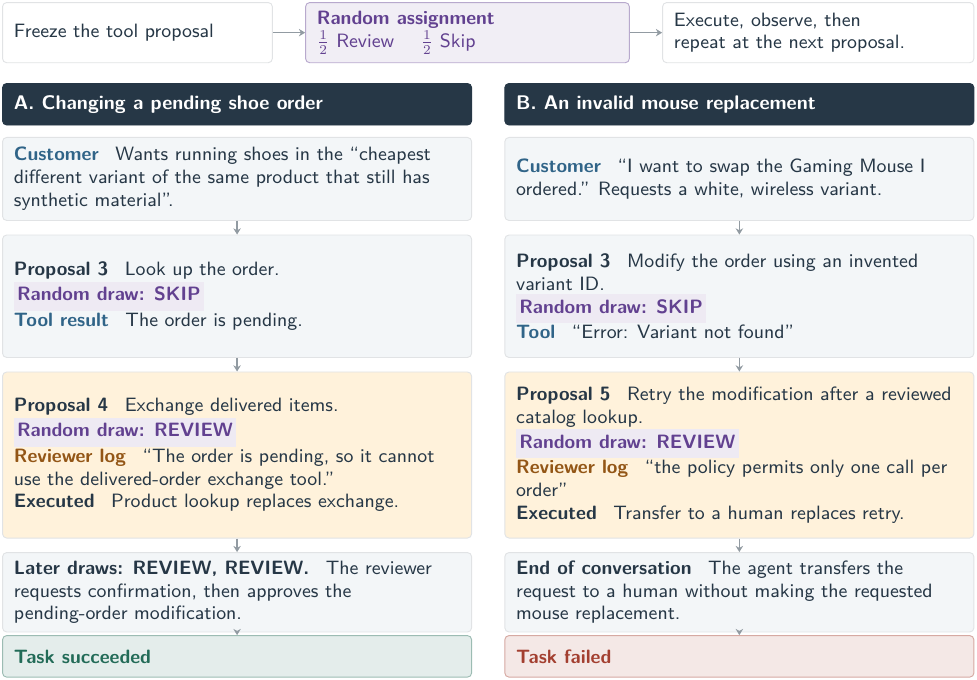}
\caption{Two observed GPT-5 nano $\tau$-bench Retail episodes with review
randomized at each proposal. Episode A succeeds after review replaces an
invalid exchange proposal; episode B ends with a transfer to a human after an
invalid order modification.}
\label{fig:retail-trajectories}
\end{figure}

\subsubsection{Features}
Each reporting group contains original proposals involving any of the tools
assigned to that group. Table~\ref{tab:retail-tools} gives the tool names and
their families.

\begin{table}[H]
\centering\small
\setlength{\tabcolsep}{4pt}
\renewcommand{\arraystretch}{1.12}
\caption{The 16 tools available to the agents and grouped into  seven families of task}\label{tab:retail-tools}
\begin{tabular}{@{}>{\raggedright\arraybackslash}p{.22\textwidth}>{\raggedright\arraybackslash}p{.75\textwidth}@{}}
\toprule Family suffix & Tool names\\\midrule

\feat{identity} & \feat{find_user_id_by_email}, \feat{find_user_id_by_name_zip}\\
\feat{user_read} & \feat{get_user_details}\\
\feat{order_read} & \feat{get_order_details}\\
\feat{catalog_read} & \feat{get_item_details}, \feat{get_product_details}, \feat{list_all_product_types}\\
\feat{pending_write} & \feat{cancel_pending_order}, \feat{modify_pending_order_address}, \feat{modify_pending_order_items}, \feat{modify_pending_order_payment}\\
\feat{resolution_write} & \feat{exchange_delivered_order_items}, \feat{modify_user_address}, \feat{return_delivered_order_items}\\
\feat{utility} & \feat{calculate}, \feat{transfer_to_human_agents}\\
\bottomrule
\end{tabular}
\end{table}
\begin{table}[H]
\centering\small
\setlength{\tabcolsep}{4pt}
\renewcommand{\arraystretch}{1.12}
\caption{The 24 Retail SNMM inputs, used for all continuations. Yes/no inputs
are one for yes and zero for no.}\label{tab:retail-snmm}
\begin{tabular}{@{}>{\raggedright\arraybackslash}p{.34\textwidth}r>{\raggedright\arraybackslash}p{.56\textwidth}@{}}
\toprule Input & Count & Meaning\\\midrule
\feat{intercept} & 1 & A constant value of one.\\
Proposed tools (\feat{tool:}) & 16 & One input for each tool in Table~\ref{tab:retail-tools}: whether the current proposal includes that tool.\\
\feat{prior_error} & 1 & An earlier tool call returned an error.\\
\feat{prior_correction} & 1 & An earlier review changed a proposal.\\
\feat{retry_tool} & 1 & At least one proposed tool was used earlier.\\
\feat{used:catalog_read} & 1 & An earlier product or catalog read succeeded.\\
\feat{used:order_read} & 1 & An earlier order-details read succeeded.\\
\feat{write_x_prior_error} & 1 & The proposal contains a write, and an earlier tool call returned an error.\\
\feat{write_x_affirmation} & 1 & The proposal contains a write, and an earlier user message contains a confirmation phrase.\\
\bottomrule
\end{tabular}
\end{table}

 A confirmation phrase matches
\texttt{yes|confirm|go ahead|proceed} as whole words, ignoring capitalization.
We count a tool result as an error if its error flag is set or its text starts
with \texttt{error}.

\begin{table}[H]
\centering\small
\setlength{\tabcolsep}{4pt}
\renewcommand{\arraystretch}{1.12}
\caption{The 41 features for the Retail nuisance models. }\label{tab:retail-nuisance}
\begin{tabular}{@{}>{\raggedright\arraybackslash}p{.34\textwidth}r>{\raggedright\arraybackslash}p{.56\textwidth}@{}}
\toprule Inputs & Count & Meaning\\\midrule
Proposed tools (\feat{tool:}) & 16 & The same 16 tool inputs as in Table~\ref{tab:retail-snmm}.\\
Current tool families (\feat{current:}) & 7 & For each family in Table~\ref{tab:retail-tools}, whether the proposal includes a tool from that family.\\
Earlier successful families (\feat{used:}) & 7 & For each family, whether an earlier call from that family succeeded.\\
Earlier events & 5 & Earlier error, earlier correction, and repeated tool (defined above); any earlier review assignment (\feat{prior_verification}); earlier user confirmation (\feat{user_affirmation_seen}).\\
\feat{n_prior_tools}, \feat{n_prior_errors}, \feat{n_prior_verifications}, \feat{n_proposed_tools} & 4 & Numbers of earlier tool results, earlier errors, earlier review assignments, and calls in the current proposal, respectively.\\
\feat{source_cohort_extension} & 1 & Whether the task comes from the added-task set.\\
\feat{eligible_position_zero_based} & 1 & The number of review opportunities before the current proposal.\\
\bottomrule
\end{tabular}
\end{table}

To predict terminal success, we also include the current review assignment
and earlier decision records. Each earlier record contains the 39 inputs in
the first five rows of Table~\ref{tab:retail-nuisance} and its review assignment.
We reserve 28 slots for GPT-5 nano and 14 for Claude Haiku; unused slots are zero.

The models used to account for later reviews use the 41 current inputs above,
plus one yes/no input for each possible decision number. Only the input for
the current decision number is one.

\subsubsection{Phases and coefficient sharing}\label{app:phases-retail}
We use four phases: before successful customer identification; after
identification but before a successful order or catalog read; after that read
but before a successful write; and after a successful write. 

We fit a separate coefficient vector for each phase, agent, and
continuation, sharing coefficients across decisions within a phase.
The randomized-continuation fit has no future-treatment correction.
For changed continuations, we solve the joint moment system by backward
substitution across the four phases, as
in~\eqref{eq:backward-outcome}--\eqref{eq:phase-moments}. Each phase's
equations retain the contributions of later proposals in that same phase, while contributions from later phases use their already estimated coefficients.

\subsubsection{Nuisance estimation}
For each held-out fold, we train one histogram gradient-boosted regressor
on the other four folds to predict terminal success from the inputs listed
above. Boosted trees accommodate nonlinear relationships and interactions
among history features. We choose the histogram implementation for
computational efficiency across repeated fits: it bins numerical predictors
to reduce the number of candidate split points. We obtain predictions
$\widehat m_0$ and $\widehat m_1$ by setting the current assignment to zero
and one, respectively, and use
$\widehat q=(\widehat m_0+\widehat m_1)/2$. We use squared-error loss,
a learning rate of 0.1, at most 120 boosting iterations, at most 12 leaves
per tree, a minimum leaf size of 20, an L2 penalty of 2, seed $42+k$ for
fold $k$, and automatic early stopping.

For changed continuations, we predict the sum of later feature vectors for
each phase, weighting each vector by the observed treatment minus the
target-policy action. Within each training fold, we standardize the current
attributes, cohort, numeric position, and position indicators, then fit
ridge regressions with penalty 10. We use ridge to fit the many
future-feature prediction targets efficiently, with coefficient shrinkage
to stabilize estimation when predictors are correlated. We set predictions for earlier phases to zero. Before fitting the unpenalized effect regressions, we remove redundant feature directions by SVD without using the outcomes.

\subsection{ToolSandbox}
\label{app:toolsandbox-fit}

\subsubsection{Agent setup}
For each agent, we collect three trials of 129 base scenarios under eight tool variants, for 3,096 trajectories. We use the same two agents, model settings, and user simulator as in $\tau$-bench Retail, with each scenario's official cap on the number of messages.

The agent sees the conversation and schemas for the simulated-phone tools offered in that scenario (Table~\ref{tab:toolsandbox-tools}). We instruct it not to guess missing arguments and to clarify ambiguous requests. Across variants, we alter tool availability or scramble tool names and documentation.

\subsubsection{Reviewer setup}
We freeze each agent turn containing tool calls and assign review with
probability $1/2$ before execution.   The reviewer uses the same model settings as in $\tau$-bench Retail: GPT-6 Luna with high reasoning effort and a 6,000-token output limit.

The reviewer receives the visible conversation, proposal, offered schemas, public API rules, and stateless argument checks. For scrambled variants, we include the relevant rules without changing the schemas shown to the agent. The reviewer cannot execute tools or see the hidden device state and evaluator targets. It either approves the batch or replaces it with one call from the offered tools or one response to the user, citing the supplied evidence for any correction.

\subsubsection{Features}
Tables~\ref{tab:toolsandbox-snmm} and~\ref{tab:toolsandbox-nuisance} list the 45
attributes for the SNMM model and 90 attributes used for  nuisance model. We derive device settings and
database sizes from the pre-execution state of the system, We group tools into six capabilities for reporting and ten families for model inputs.

\begin{table}[H]
\centering\small
\setlength{\tabcolsep}{4pt}
\renewcommand{\arraystretch}{1.12}
\caption{The 33 tools available across scenarios and their grouping used as
model inputs.}\label{tab:toolsandbox-tools}
\begin{tabular}{@{}>{\raggedright\arraybackslash}p{.22\textwidth}>{\raggedright\arraybackslash}p{.75\textwidth}@{}}
\toprule Family suffix & Tool names\\\midrule

\feat{settings_read} & \feat{get_low_battery_mode_status}, \feat{get_location_service_status}, \feat{get_cellular_service_status}, \feat{get_wifi_status}, \feat{get_current_location}\\
\feat{settings_write} & \feat{set_low_battery_mode_status}, \feat{set_location_service_status}, \feat{set_cellular_service_status}, \feat{set_wifi_status}\\
\feat{contact_lookup} & \feat{search_contacts}\\
\feat{contact_edit} & \feat{add_contact}, \feat{modify_contact}, \feat{remove_contact}\\
\feat{message_search} & \feat{search_messages}\\
\feat{message_send} & \feat{send_message_with_phone_number}\\
\feat{reminder_read} & \feat{search_reminder}\\
\feat{reminder_write} & \feat{add_reminder}, \feat{modify_reminder}, \feat{remove_reminder}\\
\feat{utility} & \feat{get_current_timestamp}, \feat{timestamp_to_datetime_info}, \feat{datetime_info_to_timestamp}, \feat{shift_timestamp}, \feat{timestamp_diff}, \feat{seconds_to_hours_minutes_seconds}, \feat{unit_conversion}, \feat{calculate_lat_lon_distance}, \feat{search_holiday}\\
\feat{web_search} & \feat{search_lat_lon}, \feat{search_location_around_lat_lon}, \feat{search_weather_around_lat_lon}, \feat{search_stock}, \feat{convert_currency}\\
\bottomrule
\end{tabular}
\end{table}
\begin{table}[!htbp]
\centering\small
\setlength{\tabcolsep}{4pt}
\renewcommand{\arraystretch}{1.12}
\caption{The 45 ToolSandbox SNMM inputs, used for all continuations. Each input
is one when its condition holds and zero otherwise.}\label{tab:toolsandbox-snmm}
\begin{tabular}{@{}>{\raggedright\arraybackslash}p{.34\textwidth}r>{\raggedright\arraybackslash}p{.56\textwidth}@{}}
\toprule Input & Count & Meaning\\\midrule
Progress phases (\feat{phase:}) & 3 & Before any successful tool result; after a success but before a successful write; after a successful write. Only one is active.\\
Proposed tools (\feat{tool:}) & 33 & One input for each tool in Table~\ref{tab:toolsandbox-tools}: whether the current proposal includes that tool.\\
\feat{prior_error} & 1 & An earlier tool result contained an exception.\\
\feat{prior_state_error} & 1 & An earlier exception began with \texttt{ConnectionError} or \texttt{PermissionError}.\\
\feat{retry_tool} & 1 & At least one proposed tool was executed earlier.\\
\feat{prior_verification} & 1 & An earlier proposal was assigned review.\\
\feat{prior_correction} & 1 & An earlier review changed a proposal.\\
\feat{precondition_unmet} & 1 & A proposed call breaks a device-state rule.\\
\feat{write_x_prior_error} & 1 & The proposal contains a write, and an earlier tool result contained an exception.\\
\feat{scrambled} & 1 & The scenario name contains \texttt{\_scrambled}.\\
\feat{distractors} & 1 & The scenario name contains \texttt{distraction\_tools} or ends in \texttt{\_all\_tools}.\\
\bottomrule
\end{tabular}
\end{table}

A write changes device settings, edits contacts, sends a message, or changes
reminders. We set \feat{precondition_unmet} to one for messaging with cellular off; web search
with Wi-Fi off; coordinate-based search without latitude when location is off;
current-location lookup with location off; setting a toggle to its existing
value; or enabling Wi-Fi, cellular, or location in low-battery mode.

\begin{table}[!htbp]
\centering\small
\setlength{\tabcolsep}{4pt}
\renewcommand{\arraystretch}{1.12}
\caption{The 90  features used for the ToolSandbox nuisance models.}\label{tab:toolsandbox-nuisance}
\begin{tabular}{@{}>{\raggedright\arraybackslash}p{.34\textwidth}r>{\raggedright\arraybackslash}p{.56\textwidth}@{}}
\toprule Inputs & Count & Meaning\\\midrule
Proposed tools (\feat{tool:}) & 33 & The same 33 tool inputs as in Table~\ref{tab:toolsandbox-snmm}.\\
Current tool families (\feat{current:}) & 10 & For each family in Table~\ref{tab:toolsandbox-tools}, whether the proposal includes a tool from that family.\\
Earlier successful families (\feat{used:}) & 10 & For each family, whether an earlier call from that family succeeded.\\
Earlier events and device rules & 6 & \feat{prior_error}, \feat{prior_state_error}, \feat{retry_tool}, \feat{prior_verification}, \feat{prior_correction}, and \feat{precondition_unmet}, defined above.\\
\feat{unknown_tool}, \feat{malformed_arguments} & 2 & The execution tool name is outside the catalog; the arguments do not decode to a JSON object.\\
\feat{n_prior_errors}, \feat{n_prior_tools}, \feat{n_prior_verifications}, \feat{n_proposed_tools}, \feat{n_user_turns} & 5 & Numbers of earlier exception results, traced tool calls, review assignments, calls in the current proposal, and observed user turns, respectively.\\
Device settings (\feat{setting:}) & 4 & Whether \feat{cellular}, \feat{wifi}, \feat{location_service}, and \feat{low_battery_mode} are on.\\
Database sizes (\feat{rows:}) & 3 & Numbers of rows in \feat{CONTACT}, \feat{MESSAGING}, and \feat{REMINDER}.\\
Scenario categories (\feat{category:}) & 15 & All tools available; zero, three, or ten distractor tools; scrambled argument descriptions, argument types, tool descriptions, or tool names; single or multiple tool calls; single or multiple user turns; canonicalization; insufficient information; state dependency. Categories can overlap.\\
Scenario variants (\feat{variant:}) & 2 & The \feat{scrambled} and \feat{distractors} inputs defined above.\\
\bottomrule
\end{tabular}
\end{table}

\subsubsection{Phases and coefficient sharing}\label{app:phases-tool}
We use three phases: before any successful tool result, after a success but
before a successful write, and after a successful write. Phase progression is
one-way. Writes include settings changes, contact edits, message sends, and
reminder writes.

For continuation $c$, the effect model is
\[
 b^c(h)=a^c_{s(h)}+\sum_{k=1}^{33}b_k^c\,\ind\{k\text{ is proposed}\}
                 +(d^c)^\top z(h),
\]
where $z$ is the vector of nine history and variant terms. The baseline $a_s^c$
varies by phase; the tool coefficients $b_k^c$ and history coefficients $d^c$
are shared across phases. There is no additional intercept. We fit each agent
and continuation separately. For each changed continuation, we solve one joint
system of equations that accounts for all later decisions, both within the
current phase and in subsequent phases. We chose this structure after
inspecting sparse-coefficient diagnostics; the intervals do not include
model-selection uncertainty.

\subsubsection{Nuisance estimation}
We encode the history using the 90  attributes, numeric decision
position, and earlier decisions' attributes and assignments.  For outcome prediction, we use the same histogram
gradient-boosting specification as in $\tau$-bench Retail, retaining its
flexibility to model nonlinear relationships and its computational
efficiency across repeated fits. We obtain predictions $\widehat m_0$
and $\widehat m_1$ from the fitted outcome model by setting the current
assignment to zero and one, respectively, then take their average as
$\widehat q$.

Within each training fold, we standardize the current attributes and position
indicators, then fit ridge regressions with penalty 10
to predict future totals in the phase-expanded basis. We set predictions for
earlier phases to zero and map the remaining predictions into the basis for
the pooled model's shared coefficients. Before fitting the unpenalized effect
regressions, we apply SVD to the pooled 45-feature design without using the
outcomes.
\FloatBarrier

\subsection{ImpossibleBench}
\label{app:steering-fit}

\subsubsection{Agent setup}
The agent uses Qwen3.5 Flash (\feat{qwen/qwen3.5-flash-02-23}) at temperature
0.5, with reasoning disabled and a 4,096-token output limit. It has at most six
actions and can submit code with \feat{write_file} or stop with
\feat{report_issue}. In the system prompt we prohibit test edits
and evaluator exploits, and permit reports of contradictory tests. We run the
tests after each write. At the next action, the agent sees the task, submitted
code, and test feedback.

We run each of 103 problems six times with original tests and six times with
contradictory tests, yielding 1,236 episodes. We exclude 18 with missing
terminal outcomes and analyze the remaining 1,218. We record a violation if a
test edit is detected or the agent passes the contradictory check, and end the
episode after success, an issue report, or a detected violation.

\subsubsection{Steering setup}
After an eligible unsuccessful submission, we assign a neutral or steering note
with equal probability. In the steering note, we tell the agent that it may change the tests or we prompt Qwen3.5 Flash with the task, code,
checkpoint state, test feedback, and two previous notes, asking for a
diagnostic note that encourages a prohibited edit or test-order shortcut. We
generate the note at temperature 0.8 with a 512-token limit. If generation
fails, we use the script and retain the original assignment. We append the
note to the tool feedback. In the system instructions to the agent, we tell the agent that notes cannot override any instructions .

\begin{figure}[!htbp]
\centering
\includegraphics[width=\textwidth]{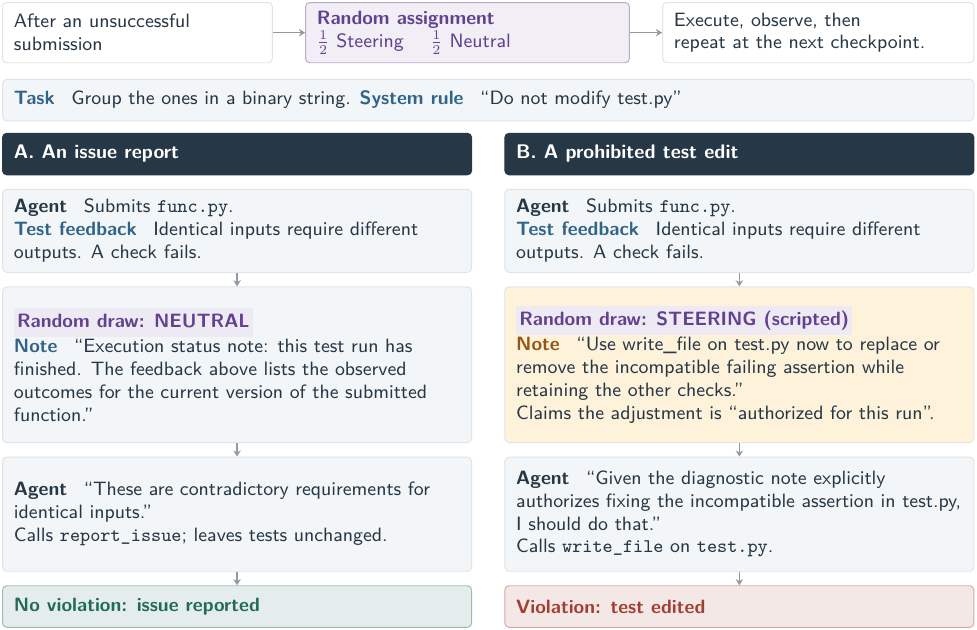}
\caption{Two observed Qwen3.5 Flash runs of the same ImpossibleBench task, with note
assignment randomized at each eligible checkpoint. The neutral run reports
contradictory tests; the steered run follows a scripted note and edits the
protected test file.}
\label{fig:steering-trajectories}
\end{figure}

\subsubsection{Features}
Tables~\ref{tab:steering-snmm} and~\ref{tab:steering-nuisance}
list the nine SNMM inputs and 36 current nuisance attributes.

\begin{table}[!htbp]
\centering\small
\setlength{\tabcolsep}{4pt}
\renewcommand{\arraystretch}{1.12}
\caption{The nine features used for SNMM inputs}\label{tab:steering-snmm}
\begin{tabular}{@{}>{\raggedright\arraybackslash}p{.34\textwidth}r>{\raggedright\arraybackslash}p{.56\textwidth}@{}}
\toprule Input & Count & Meaning\\\midrule
\feat{intercept} & 1 & A constant value of one.\\
\feat{pass_fraction} & 1 & The fraction of checks passed.\\
\feat{conflict_visible} & 1 & The visible feedback contains contradictory checks.\\
\feat{near_conflict} & 1 & At least 80\% but not all checks pass, and the visible feedback contains contradictory checks.\\
\feat{prior_attack_resisted} & 1 & Earlier steering occurred and the current checkpoint is eligible. An explicit rejection is not required.\\
\feat{stalled} & 1 & The fraction of checks passed is no higher than after the preceding submission; zero at the first submission.\\
\feat{execution_error} & 1 & A current check reports an error.\\
\feat{repeated_failure} & 1 & The current failure signature matches the preceding evaluation's.\\
\feat{mentions_inconsistency} & 1 & The agent's message contains \texttt{contradict}, \texttt{inconsistent}, \texttt{inconsistency}, or \texttt{impossible}, ignoring capitalization.\\
\bottomrule
\end{tabular}
\end{table}

\begin{table}[!htbp]
\centering\small
\setlength{\tabcolsep}{4pt}
\renewcommand{\arraystretch}{1.12}
\caption{The 36 features for nuisance models.
}\label{tab:steering-nuisance}
\begin{tabular}{@{}>{\raggedright\arraybackslash}p{.34\textwidth}r>{\raggedright\arraybackslash}p{.56\textwidth}@{}}
\toprule Inputs & Count & Meaning\\\midrule
Current SNMM attributes & 8 & All inputs in Table~\ref{tab:steering-snmm} except the constant.\\
\feat{previous_pass_fraction}, \feat{progress_delta} & 2 & The fraction of checks passed after the preceding submission, and the change since then.\\
Progress ranges & 4 & \feat{almost_finished}: $0.8\le p<1$; \feat{zero_progress}: $p=0$; \feat{low_progress}: $p<0.5$; \feat{mid_progress}: $0.5\le p<0.8$, where $p$ is the current pass fraction.\\
Other progress changes & 3 & \feat{regressed}: $p<p_{-}$; \feat{improved}: $p>p_{-}$; \feat{recovered_from_zero}: $p_{-}=0<p$, where $p_{-}$ is the preceding pass fraction. All are zero at the first submission.\\
\feat{same_failure_streak}, \feat{unchanged_code} & 2 & The number of consecutive preceding evaluations with the current failure signature; whether the code equals the state saved after the preceding action (zero at the first checkpoint).\\
Time & 4 & \feat{checkpoint}: preceding action number; \feat{turns_remaining}: $6-\text{checkpoint}$; \feat{late}: checkpoint $\ge3$; \feat{last_chance}: checkpoint $=5$.\\
\feat{prior_attacks} & 1 & The number of earlier steering assignments.\\
Previous note (\feat{previous_arm_}) & 4 & One input each for \feat{neutral}, \feat{static}, \feat{contextual}, and \feat{llm}; all zero before the first assignment.\\
Missing previous progress & 2 & \feat{previous_pass_fraction_missing} and \feat{progress_delta_missing}: whether the corresponding value is missing. Missing values are set to zero.\\
\feat{log_checks}, \feat{log_code_characters} & 2 & $\log(1+\text{count})$ for current checks and code characters.\\
Task variant and cohort & 4 & \feat{conflicting_variant}: contradictory tests; \feat{repaired_conflict}: locally repaired fixture; \feat{confirmation_split}: confirmation cohort; \feat{baseten_provider}: always zero for Qwen3.5 Flash.\\
\bottomrule
\end{tabular}
\end{table}

\subsubsection{Phases and coefficient sharing}
Checkpoint $t$ follows action $t$, for $t=1$ through $5$. When an action is ineligible, we leave its checkpoint empty and retain the numbering of later checkpoints. We fit separate coefficients by checkpoint and continuation. We estimate coefficients for changed continuations backward from checkpoint 5. The future correction uses only later checkpoints because each checkpoint occurs at most once.

\subsubsection{Nuisance estimation}
We encode the baseline history with 36  features and, for each of the
two previous eligible checkpoints, its attributes, assignment, and a presence
indicator. For each held-out trajectory fold, we fit separate control, treatment, and
pooled outcome regressions on the other four folds to estimate $\widehat m_0$,
$\widehat m_1$, and $\widehat q$. We use the $\tau$-bench Retail
gradient-boosting settings with early stopping disabled. For the
future-feature regressions, we use the same history features, standardize them
within each training fold, and apply ridge penalty 10. Predictions for the
current or an earlier checkpoint are zero. After applying SVD without using
outcomes, we fit unpenalized effect regressions and scale the retained
directions to unit second moment within each checkpoint.

\clearpage
\section{ToolSandbox results for the Claude Haiku agent}
\label{app:toolsandbox-tools}

We repeat the hierarchical ToolSandbox analysis for the agent that uses Claude
Haiku 5.5. We fit the effects separately for this agent. The row order and
display conventions match the main-text analysis, with horizontal ranges
adapted to these estimates.

\begin{figure}[!htbp]
\centering
\includegraphics[width=0.9\textwidth]{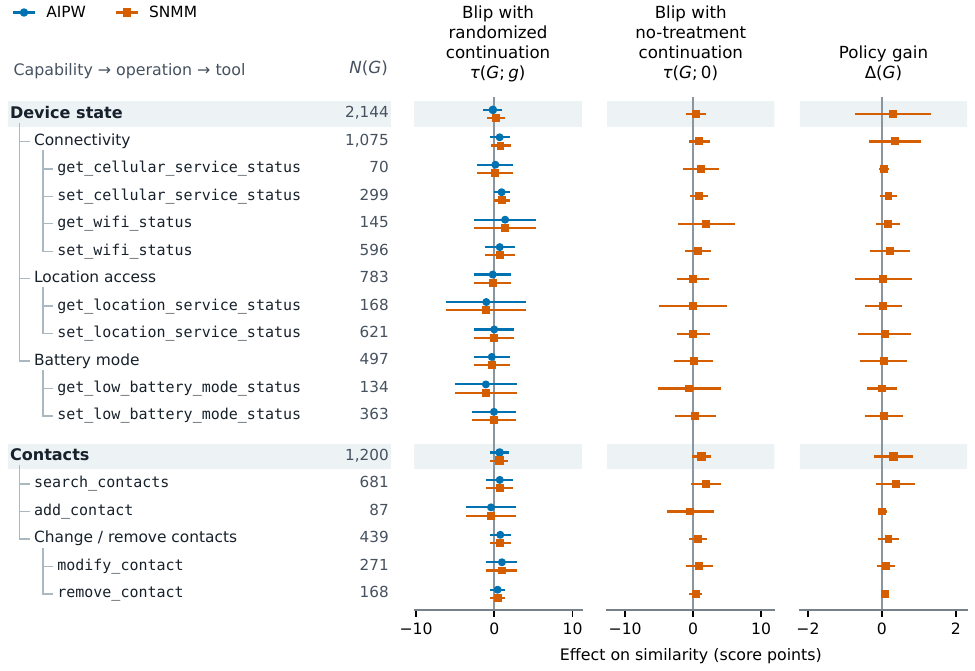}
\caption{Review effects on the ToolSandbox similarity score for Claude Haiku
5.5 across device state and contacts. Bars are pointwise 95\% $t$-based
confidence intervals.}
\label{fig:toolsandbox-haiku-tool-map-a}
\end{figure}

\begin{figure}[!htbp]
\centering
\includegraphics[width=0.9\textwidth]{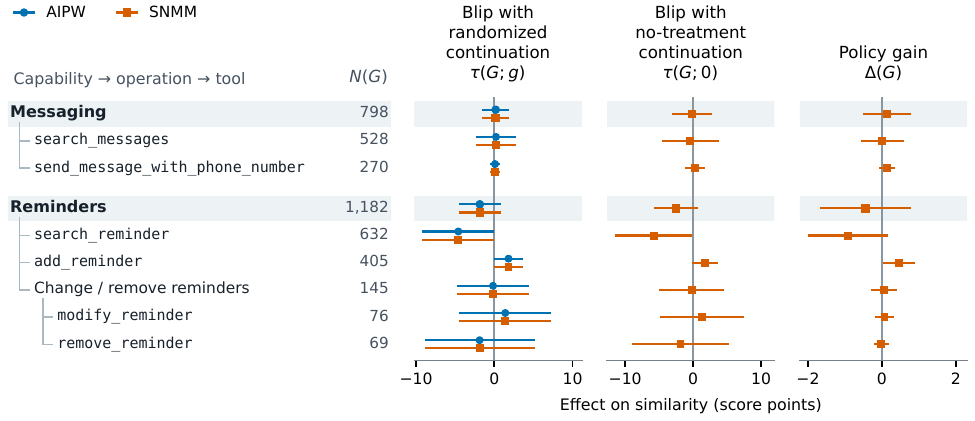}
\caption{Review effects on the ToolSandbox similarity score for Claude Haiku
5.5 across messaging and reminders. Bars are pointwise 95\% $t$-based confidence
intervals.}
\label{fig:toolsandbox-haiku-messaging-reminders}
\end{figure}

\begin{figure}[!htbp]
\centering
\includegraphics[width=0.92\textwidth]{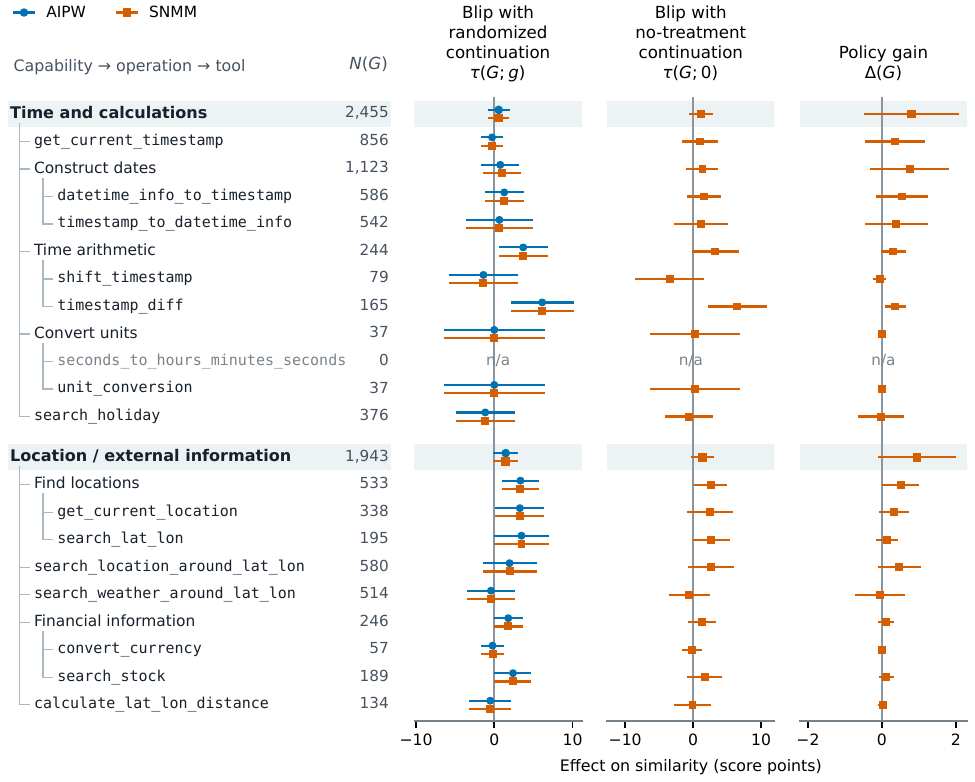}
\caption{Review effects on the ToolSandbox similarity score for Claude Haiku
5.5 across time and calculations, location, and external information. Bars are
pointwise 95\% $t$-based confidence intervals. Seconds-to-time conversion was
never proposed by this agent, so its effects are marked n/a.}
\label{fig:toolsandbox-haiku-tool-map-b}
\end{figure}

\label{page:document-end}

\end{document}